\documentclass[5p,times]{elsarticle}

\usepackage{booktabs}
\usepackage{graphicx}
\usepackage{wrapfig}
\usepackage{mwe}
\usepackage[table]{xcolor}

\usepackage{url}

\usepackage{breakurl}
\usepackage[colorlinks,hidelinks]{hyperref}
\usepackage[capitalize,noabbrev,nameinlink]{cleveref}

\usepackage{array}
\newcolumntype{L}[1]{>{\raggedright\let\newline\\\arraybackslash\hspace{0pt}}m{#1}}
\newcolumntype{C}[1]{>{\centering\let\newline\\\arraybackslash\hspace{0pt}}m{#1}}
\newcolumntype{D}[1]{>{\centering\let\newline\\\arraybackslash}m{#1}}
\newcolumntype{R}[1]{>{\raggedleft\let\newline\\\arraybackslash\hspace{0pt}}m{#1}}

\makeatletter
\let\@afterindenttrue\@afterindentfalse
\makeatother

\usepackage{lineno}
\modulolinenumbers[5]

\journal{Big Data Research - Elsevier}

\begin{document}

\begin{frontmatter}

\title{Data Science Methodologies: \\ Current Challenges and Future Approaches}

\author[VICOMTECH,TECNUN]{I\~nigo Martinez\corref{cor1}}
\ead{imartinez@vicomtech.org}

\author[TECNUN,ICDIA]{Elisabeth Viles}
\ead{eviles@tecnun.es}

\author[VICOMTECH]{Igor G Olaizola}
\ead{iolaizola@vicomtech.org}

\cortext[cor1]{Corresponding author}

\address[VICOMTECH]{Vicomtech Foundation, Basque Research and Technology Alliance (BRTA), Donostia-San Sebastián 20009, Spain}

\address[TECNUN]{TECNUN School of Engineering, University of Navarra, Donostia-San Sebastián 20018, Spain}

\address[ICDIA]{Institute of Data Science and Artificial Intelligence, University of Navarra, Pamplona 31009, Spain}

\begin{abstract}
	Data science has employed great research efforts in developing advanced analytics, improving data models and cultivating new algorithms. However, not many authors have come across the organizational and socio-technical challenges that arise when executing a data science project: lack of vision and clear objectives, a biased emphasis on technical issues, a low level of maturity for ad-hoc projects and the ambiguity of roles in data science are among these challenges. Few methodologies have been proposed on the literature that tackle these type of challenges, some of them date back to the mid-1990, and consequently they are not updated to the current paradigm and the latest developments in big data and machine learning technologies. In addition, fewer methodologies offer a complete guideline across team, project and data \& information management. In this article we would like to explore the necessity of developing a more holistic approach for carrying out data science projects. We first review methodologies that have been presented on the literature to work on data science projects and classify them according to the their focus: project, team, data and information management. Finally, we propose a conceptual framework containing general characteristics that a methodology for managing data science projects with a holistic point of view should have. This framework can be used by other researchers as a roadmap for the design of new data science methodologies or the updating of existing ones.
\end{abstract}


\end{frontmatter}


\section{Introduction} \label{introduction}

In recent years, the field of data science has received increased attention and has employed great research efforts in developing advanced analytics, improving data models and cultivating new algorithms. The latest achievements in the field are a good reflection of such undertaking \cite{state_of_ai}. In fact, the data science research community is growing day by day, exploring new domains, creating new specialized and expert roles, and sprouting more and more branches of a flourishing tree called data science. Besides, the data science tree is not alone, as it is nourished by the neighboring fields of mathematics, statistics and computer science. 

However, these recent technical achievements do not go hand in hand with their application to real data science projects.
In 2019, VentureBeat \cite{venturebeat_ds_project} revealed that 87\% of data science projects never make it into production and a NewVantage survey \cite{new_vantage_summary} reported that for 77\% of businesses the adoption of big data and artificial intelligence (AI) initiatives continue to represent a big challenge. Also  \cite{gartner_advanced_analytics} reported that 80\% of analytics insights will not deliver business outcomes through 2022 and 80\% of data science projects will ``remain alchemy, run by wizards'' through 2020. With the competitive edge that such advanced techniques provide to researchers and practitioners, it is quite remarkable to witness such low success rates in data science projects.

Regarding how data science teams are approaching and developing projects across diverse domains, Leo Breiman \cite{breiman_cultures} states that there are two cultures on statistical modeling: a) the forecasting branch which is focused on building efficient algorithms for getting good predictive models to forecast the future, and b) the modeling branch that is more interested on understanding the real world and the underlying processes. From the latter perspective, theory, experience, domain knowledge and causation really matter, and the scientific method is essential. On this wise, philosophers of science from the XX century such as Karl Popper, Thoman Kuhn, Imre Lakatos, or Paul Feyerabend theorized about how the process of building knowledge and science unfolds \cite{garcia_epistemology}. Among other things, these authors discussed the nature and derivation of scientific ideas, the formulation and use of the scientific method, and the implications of the different methods and models of science. In spite of the obvious differences, in this article we would like to motivate data scientists to discuss the formulation and use of the scientific method for data science research activities, along the implications of different methods for executing industry and business projects. At present, data science is a young field and conveys the impression of a handcrafted work. However, among all the uncertainty and the exploratory nature of data science, there could be indeed a rigorous ``science'', as it was understood by the philosophers of science, and that can be controlled and managed in an effective way.

On the literature, few authors \cite{saltz_new_processes, saltz_sociotechnical_challenges} have come across the organizational and socio-technical challenges that arise when executing a data science project, for instance: lack of vision, strategy and clear objectives, a biased emphasis on technical issues, lack of reproducibility, and the ambiguity of roles are among these challenges, which lead to a low level of maturity in data science projects, that are managed in ad-hoc fashion.

Even though these issues do exist in real-world data science projects, the community has not been overly concerned about them, and not enough has been written about the solutions to tackle these problems. As it will be illustrated on \cref{results}, some authors have proposed methodologies to manage data science projects, and have come up with new tools and processes to handle the cited issues.

However, while the proposed solutions are leading the way to tackle these problems, the reality is that data science projects are not taking advantage of such methodologies. On a survey \cite{saltz_exploring_pm} carried out in 2018 to professionals from both industry and not-for-profit organizations, 82\% of the respondents did not follow an explicit process methodology for developing data science projects, but 85\% of the respondents considered that using an improved and more consistent process would produce more effective data science projects. 

Considering a survey from KDnuggets in 2014 \cite{piatetsky_crispdm}, the main methodology used by 43\% of responders was CRISP-DM. This methodology has been consistently the most commonly used for analytics, data mining and data science projects, for each KDnuggets poll starting in 2002 up through the most recent 2014 poll \cite{saltz_crispdm}. Despite its popularity, CRISP-DM was created back in the mid-1990 and has not been revised since.

Therefore, the aim of this paper is to conduct a critical review of methodologies that help in managing data science projects, classifying them according to their focus and evaluating their competences dealing with the existing challenges. As a result of this study we propose a conceptual framework containing features that a methodology for managing data science projects with a holistic point of view could have. This scheme can be used by other researchers as a roadmap to expand currently used methodologies or to design new ones.

From this point onwards, the paper is structured as follows:
A contextualization of the problem is presented in \cref{theoretical_framework}, where some definitions on the terms ``data science'' and ``big data'' are provided to avoid any semantic misunderstanding and additionally frame what a data science project is about.  \cref{theoretical_framework} also presents the organizational and socio-technical challenges that arise when executing a data science project.
\cref{research_methodology} describes the research methodology used in the article and introduces the research questions.
\cref{results} presents a critical review of data science project methodologies.
\cref{discussion} discusses the results obtained and finally, \cref{conclusions} describes the directions on how to extend this research in the future and the main conclusions of the paper.

\section{Theoretical Framework}  \label{theoretical_framework}
\subsection{Background and definitions}  \label{background_definition}

In order to avoid any semantic misunderstanding, we start by defining the term ``data science'' more precisely. Its definition will help explaining the particularities of data science projects and frame the management methodology proposed in this paper.

\subsubsection{Data science}

Among the authors that assemble data science from long-established areas of sciences, there is a overall agreement on the fields that feed and grow the tree of data science.
For instance, \cite{granville_data_scientist} defines data science as the intersection of computer science, business engineering, statistics, data mining, machine learning, operations research, six sigma, automation and domain expertise,
whereas \cite{frank_lo_data_science} states that data science is a multidisciplinary intersection of mathematics expertise, business acumen and hacking skills.
For \cite{loukides_data_science}, data science requires skills ranging from traditional computer science to mathematics to art
and \cite{conway_venn_diagram} presents a Venn diagram with data science visualized as the joining of a) hacking skills b) math and stats knowledge and c) substantive expertise.
By contrast, for the authors of \cite{jones_statistical_engineering}, many data science problems are statistical engineering problems, but with larger, more complex data that may require distributed computing and machine learning methods in addition to statistical modeling.

Yet, few authors look on the fundamental purpose of data science, which is crucial to understand the role of data science in the business \& industry areas and its possible domain applications.
For the authors of \cite{saltz_exploring_pm}, data science is the analysis of data to solve problems and develop insights, whereas \cite{das_systematic_research} states that data science uses ``statistical and machine learning techniques on big multi-structured data in a distributed computing environment to identify correlations and causal relationships, classify and predict events, identify patterns and anomalies, and infer probabilities, interest and sentiment''. For them, data science combines expertise across software development, data management and statistics.
As on \cite{saltz_ds_roles}, data science is described as the field that studies the computational principles, methods and systems for extracting and structuring knowledge from data.

There has also been a growing interest on defining the work carried out by data scientists and listing the necessary skills to become a data scientist, in order to better understand its role among traditional job positions.
In this sense, \cite{wills_tweet} defines data scientist as a person who is ``better at statistics than any software engineer and better at software engineering than any statistician'' as 
\cite{warden_data_science} presents a ``unicorn'' perspective of data scientists, and states that data scientists handle everything from finding data, processing it at scale, visualizing it and writing up as a story.

Among the hundreds of different interpretations that can be found for data science, we take the comments included above as a reference and provide our own definition to frame the rest of the article:

\textit{Data science is an multidisciplinary field that lies between computer science, mathematics and statistics, and comprises the use of scientific methods and techniques, to extract knowledge and value from large amounts of structured and/or unstructured data.}

Hence, from this definition we draw that data science projects aim at solving complex real problems via data-driven techniques. In this sense, data science can be applied to almost every existing sector and domain out there: 
banking (fraud detection \cite{fraud_detection},
credit risk modeling \cite{risk_modelling},
customer lifetime value \cite{customer_value}), 
finance (customer segmentation \cite{customer_segmentation},
risk analysis \cite{risk_analysis},
algorithmic trading \cite{algorithmic_trading}), 
health-care (medical image analysis \cite{image_analysis},
drug discovery \cite{drug_discovery},
bio-informatics \cite{bio_informatics},), 
manufacturing optimization (failure prediction \cite{failure_prediction},
maintenance scheduling \cite{maintenance_scheduling},
anomaly detection \cite{anomaly_detection}),
e-commerce (targeted advertising \cite{targeted_advertising},
product recommendation \cite{product_recommendation},
sentiment analysis \cite{sentiment_analysis}), 
transportation (self driving cars \cite{self_driving_cars},
supply chain management \cite{supply_chain},
congestion control \cite{congestion_control}) 
just to mention some of them. Even though data science can be seen as an application-agnostic discipline, we believe that it in order to appropriately extract value from data it is highly recommended to have an expertise in the application domain. 

\subsubsection{Big data technologies}
Looking back at the evolution of data science during the last decade, its rapid expansion is closely linked to the growing ability to collect, store and analyze data generated at an increasing frequency \cite{green_advancements_2019}. In fact, during the mid-2000s there were some fundamental changes in each of these stages (collection, storage and analysis) that shifted the paradigm of data science and big data.

With respect to collection, the growth of affordable and reliable interconnected sensors, built in smart-phones and industrial machinery, has significantly changed the way statistical analysis is approached. In fact, traditionally the acquisition cost was so high that statisticians carefully collected data in order to be necessary and sufficient to answer a specific question. This major shift in data collection has indeed provoked an explosion of the amount of machine-generated data. In regards to storage, we must highlight the development of new techniques to distribute data across nodes in a cluster and the development of distributed computation to run in parallel against data on those cluster nodes. Hadoop and Apache Spark ecosystems are good examples of new technologies that contributed to the advances in collection and storage techniques. Besides, a major breakthrough for the emergence of new algorithms and techniques for data analysis has been the increase in the computation power, both in CPUs and GPUs. Specially the latest advances in GPUs have pushed forward the expansion of deep learning techniques, very eager of fast matrix operations \cite{shi_benchmarking}. 

In addition to the improvements on collection, storage and analysis, data science has benefited from a huge community of developers and researchers, coming from cutting edge companies and research centers. In our terminology, ``big data'', which is commonly defined by the 5 V's (volume, velocity, variety, veracity, value) is a subset field within data science that focuses on data distribution and parallel processing.

Overall, during the last years the data science community has pursued the excellence and has employed great research efforts in developing advanced analytics, focusing of solving technical problems and as a consequence the organizational and socio-technical challenges have been put aside. The following section summarizes the main problems faced by data science professionals during real business and industry projects.

\subsection{Current Challenges}  \label{current_challenges}

Leveraging data science within a business organizational context involves additional challenges beyond the analytical ones. The studies mentioned at the introduction of this paper just reflect the existing difficulty in executing data science and big data projects. Here below we have gathered some of the main challenges and pain points that come out during a data science project, both at the organizational level and a technical level. 

\subsubsection*{\textbf{Coordination, collaboration and communication}}

The field of data science is evolving from a work done by individual, ``lone wolf'' data scientists, towards a work carried out by a team with specialized disciplines.  Considering data science projects as a complex team effort, \cite{jacob_spoelstra_ds_process, saltz_exploring_pm, espinosa_coordination_governance} bring up \textit{coordination}, defined as “the management of dependencies among task activities” as the biggest challenge for data science projects. Poorly coordinated processes result in confusion, inefficiencies and errors. Moreover, this lack of efficient coordination happens both within data analytics teams and across the organization \cite{colas_data_conundrum}.

Apart from lack of coordination, for \cite{stef_caraguel_ds_challenges, maguire_data_analytics, saltz_bigdata_methodologies} there are clear \textit{issues of collaboration} and \cite{kuhn_analytics_canvas,becker_bigdata_dynamics} highlight a \textit{lack of transparent communication} between the three main stakeholders: business (client), analytics team and the IT department. For instance, \cite{stef_caraguel_ds_challenges} mentions the difficulty for analytics teams to deploy to production, \textit{coordinate with the IT department} and explain data science to business partners. \cite{stef_caraguel_ds_challenges} also reveals the lack of support from the business side, in a way that there is not enough business input or domain expertise information to achieve good results. Overall, it seems that the data analytics team and data scientists in general are struggling to work efficiently alongside the IT department and the business agents.

In addition, \cite{becker_bigdata_dynamics} points out \textit{ineffective governance models} for data analytics and \cite{colas_data_conundrum} emphasize inadequate management and a lack of sponsorship from top management side. In this context, \cite{saltz_shortcomings} affirm that working in confusing, chaotic environments can be frustrating and may lower team members' motivation and their ability to focus on the project objectives.

\subsubsection*{\textbf{Building data analytics teams}}

In other terms, \cite{saltz_bigdata_projects} brings out problems to engage the proper team for the project and \cite{maguire_data_analytics, saltz_bigdata_methodologies, colas_data_conundrum, becker_bigdata_dynamics, prendki_lessons_agile} highlight the \textit{lack of people with analytics skills}. These shortages of specialized analytical labor have caused every major university to launch new big data, analytics or data science programs \cite{espinosa_coordination_governance}. In this regard, \cite{saltz_bigdata_methodologies} advocates the \textit{need for a multidisciplinary team}: data science, technology, business and management skills are necessary to achieve success in data science projects. For instance, \cite{saltz_exploring_pm} states that data science teams have a strong dependence on the leading data scientist, which is due to process immaturity and the lack of a robust team-based methodology.

\subsubsection*{\textbf{Defining the data science project}}

Data science projects often have highly uncertain inputs as well as highly uncertain outcomes, and are often ad-hoc \cite{bhardwaj_datahub}, featuring significant back-and-forth among the members of the team, and trial-and-error to identify the right analysis tools, programs, and parameters. 

The exploratory nature of such projects makes \textit{challenging to set adequate expectations} \cite{das_systematic_research}, establish realistic project timelines and estimate how long projects would take to complete \cite{saltz_sociotechnical_challenges}. In this regard, \cite{saltz_bigdata_projects, byrne_development_workflows} point out that the scope of the project can be difficult to know ex ante, and understanding the business goals is also a troubling task.

More explicitly, the authors in \cite{kuhn_analytics_canvas, colas_data_conundrum, becker_bigdata_dynamics} highlight the \textit{absence of clear business objectives}, insufficient ROI or business cases, and an improper scope of the project. For \cite{ransbotham_analytics_gap} there is a \textit{biased emphasis on the technical issues}, which has limited the ability of organizations to unleash the full potential of data analytics. Rather than focusing of the business problem, data scientists have been often obsessed with achieving state of the art results on benchmarking tasks, but searching for a small increase in performance can in fact make models too complex to be useful. This mindset is convenient for data science competitions, such as Kaggle \cite{kaggle}, but not for the industry. Kaggle competitions are actually great for machine learning education, but they can set wrong expectations about what to demand in real-life business settings \cite{jung_kaggle_competition}.

\subsubsection*{\textbf{Stakeholders vs Analytics}}

Besides, more often than not the project proposal is not clearly defined \cite{stef_caraguel_ds_challenges} and there is \textit{not enough involvement by the business side}, who might just provide the data and barely some domain information, assuming that the data analytics team will do the rest of the ``magic'' by itself. The high expectations set up by machine learning and deep learning techniques has induced a misleading perception that these new technologies can achieve whatever the business suggests at a very low cost, and this is very far from reality \cite{vanauer_bigdata_organizations}. The lack of involvement by the business side can also be caused by a lack of understanding between both parties: data scientists may not understand the domain of the data, and the business is usually not familiar with data analysis techniques. In fact, the presence of an intermediary that understands both the language of data analytics and the domain of application can be crucial to make these two parties understand each other and reduce this data science gap \cite{colson_ds_generalists}.

The exposed project management issues may be the result of a \textit{low level of process maturity} \cite{bhardwaj_datahub} and a lack of consistent methods and processes to approach the topic of data science \cite{maguire_data_analytics}. Furthermore, the consequences of such low adoption of processes and methodologies may also lead to delivering the ``wrong thing'' \cite{jacob_spoelstra_ds_process, saltz_exploring_pm, domino_data_lab}, and ``scope creep'' \cite{jacob_spoelstra_ds_process, saltz_exploring_pm}. In fact, the lack of effective processes to engage with stakeholders increases the risk that teams will deliver something that does not satisfy stakeholder needs. The most obvious portrayal of such problem is the zero impact and \textit{lack of use of the project results} by the business or the client \cite{stef_caraguel_ds_challenges}.

\subsubsection*{\textbf{Driving with data}}

The use of a data-driven approach defines the main particularity of a data science project. Data is at the epicenter of the entire project. Yet, it also produces some particular issues discussed below. Hereafter we have gathered the main challenges that arise when working with data, whether related to the tools, to the technology itself, or to information management.

A reiterated complain in real data science projects by data scientists is the \textit{quality of the data}: whether data is difficult to access \cite{saltz_bigdata_methodologies}, or whether is ``dirty'' and has issues, data scientists usually conclude that data has not enough potential to be suitable for machine learning algorithms. Understanding what data might be available \cite{saltz_bigdata_projects}, its representativeness for the problem at hand \cite{saltz_bigdata_methodologies} and its limitations \cite{byrne_development_workflows} is critical for the success of the project. In fact, \cite{domino_data_lab} states that the lack of coordinated data cleaning or \textit{quality assurance checks} can lead to erroneous results. In this regard, data scientists tend to forget about the validation stage. In order to assure a robust validation of the proposed solution under real industrial and business conditions, data and/or domain expertise must be gathered with enough anticipation.

Considering the big data perspective is also important \cite{sivarajah_bigdata_challenges}. An increase in data volume and velocity intensifies the computation requirements and hence the \textit{dependence of the project on IT resources} \cite{saltz_sociotechnical_challenges}. In addition, the scale of the data magnifies the technology complexity and the necessary architecture and infrastructure \cite{becker_bigdata_dynamics} and as a result the corresponding costs \cite{colas_data_conundrum}.

In other terms, \cite{saltz_bigdata_methodologies, colas_data_conundrum, sivarajah_bigdata_challenges} also highlight the importance of \textit{data security and privacy}, and \cite{colas_data_conundrum, becker_bigdata_dynamics} point out the complex dependency on legacy systems and data integration issues.

In relation to the limitations of machine learning algorithms, one of most repeated issues is that popular deep learning techniques require many relevant training data and their robustness is every so often called into question. \cite{prendki_lessons_agile} brings up the \textit{excessive model training and retraining costs}. In fact, data scientists tend to use 4 times as much data as they need to train machine learning models, which is resource intensive and costly.
Apart from that, \cite{prendki_lessons_agile} points out that data scientists can often focus on the \textit{wrong model performance metrics}, without making any reference to overall business goals or trade-offs between different goals. 

\begin{table*}[!htb]
	\centering
	\resizebox{1\textwidth}{!}{
	\begin{tabular}{@{}ccc@{}}
		\toprule
		\textbf{Team Management}                & \textbf{Project Management}                   & \textbf{Data \& Information Management} \\ \midrule
		Poor coordination                       & Low level of process maturity                 & Lack of reproducibility                 \\
		Collaboration issues across teams       & Uncertain business objectives                 & Retaining and accumulation of knowledge \\
		Lack of transparent communication       & Setting adequate expectations                 & Low data quality for ML                 \\
		Inefficient governance models           & Hard to establish realistic project timelines & Lack of quality assurance checks        \\
		Lack of people with analytics skills    & Biased emphasis on technical issues           & No validation data                      \\
		Rely not only on leading data scientist & Delivering the wrong thing                  & Data security and privacy               \\
		Build multidisciplinary teams           & Project not used by business                  & Investment in IT infrastructure         \\ \bottomrule
	\end{tabular}
	}
	\caption{Data science projects main challenges}
	\label{tab:challenges}
\end{table*}

\subsubsection*{\textbf{Deliver insights}}

\cite{jacob_spoelstra_ds_process, saltz_exploring_pm, sivarajah_bigdata_challenges} point out the issue of slow information and data sharing among the team members. They state that these poor processes for storing, retrieving and sharing data and documents wastes time as people need to look for information and increases the risk for using the wrong version.
In relation to this, \cite{jacob_spoelstra_ds_process, saltz_exploring_pm, domino_data_lab} reveal a \textit{lack of reproducibility} in data science projects. In fact, they call for action and development of new tools to tackle the lack of reproducibility, since it might be ``impossible to further build on past projects given the inconsistent preservation of relevant artifacts'' like data, packages, documentation, and intermediate results.

This can be a serious problem for the long-term sustainability of data science projects. In several applied data science projects, the main outcome of the project may not be the machine learning model or the predicted quantity of interest, but an intangible such as the project process itself or the generated knowledge along its development. While it is essential to reach the project goals, in some cases it is more important to know how the project did reach those goals, what path did it follow and understand why it took those steps and not other ones. This generated knowledge about the route that a data science project follows is crucial for understanding the results and paves the way for future projects. That is why this \textit{knowledge has to be managed} and preserved in good condition, and for that the ability to reproduce data science tasks and experiments is decisive. \cite{prendki_lessons_agile} states that retaining institutional knowledge is a challenge, since data scientists and developers are in short supply and may jump to new jobs. 

To tackle this problem, \cite{prendki_lessons_agile} proposes to document everything and create a detailed register for all new machine learning models, thus enabling future hires to quickly replicate work done by their predecessors. In this regard, \cite{byrne_development_workflows} noted knowledge-sharing within data science teams and across the entire organization as one key factor for project success and \cite{colas_data_conundrum, becker_bigdata_dynamics} added data \& information management as well. 

In relation to data management, \cite{prendki_lessons_agile} also pointed out the issue of multiple similar but inconsistent data sets, where many versions of the same data sets could be circulating within the company, with no way to identify which is the correct one.

\subsubsection*{\textbf{Summary}}

The presented problematic points have been classified in three main categories, according to whether they relate to the a) team or organization, b) to the project management or c) to the data \& information management. 
This taxonomy is meant to facilitate and better understand the types of problems that arise during a data science project. In addition, this classification will go hand in hand with the review of data science methodologies that will be introduced later in the paper. An alternative classification system for big data challenges is proposed on \cite{sivarajah_bigdata_challenges}, defined by data, process and management challenges. For them, data challenges are related to the characteristics of data itself, process challenges arise while processing the data and management challenges tackle privacy, security, governance and lack of skills. We claim that the proposed taxonomy of challenges incorporates this classification and has a broader view. \cref{tab:challenges} summarizes the main challenges that come out when executing real data science projects.

Some of the challenges listed on \cref{tab:challenges} are considered to be a symptom or reflection of a larger problem, which is the lack of a coherent methodology in data science projects, as was stated by \cite{saltz_new_processes}. In this sense, \cite{saltz_exploring_pm} suggested that an augmented data science methodology could improve the success rate of data science projects. In the same article, the author presented a survey carried out in 2018 to professionals from both industry and not-for-profit organizations, where 82\% of the respondents declared that they did not follow an explicit process methodology for developing data science projects, but 85\% of the respondents considered that using an improved and more consistent process would produce more effective data science projects.

Therefore, in this article we would like to explore the following research questions: 
\begin{itemize}
	
	\item \textbf{RQ1}: What methodologies can be found on the literature to manage data science projects?
	
	\item \textbf{RQ2}: Are these available methodologies prepared to meet the demands of current challenges?
	
\end{itemize}

\section{Research Methodology}   \label{research_methodology}

In order to investigate the state-of-the-art in data science methodologies, in this article we have opted for a critical review of the literature. Saunders and Rojon \cite{saunders_critical_literature} define a critical literature review as a “combination of our knowledge and understanding of what has been written, our evaluation and judgment skills and our ability to structure these clearly and logically in writing”. 
They also point out several key attributes of a critical literature review:
a) it discusses and evaluates the most relevant research relevant to the topic, 
b) it acknowledges the most important and relevant theories, frameworks and experts within the field, 
c) it contextualizes and justifies aims and objectives and
d) it identifies gaps in knowledge that has not been explored in previous literature.

The presented critical literature review was carried out based on the preceding concepts and through a comparison of literature on data science project management. The main reason to go for a critical review rather than a systematic review was that information regarding the use of data science methodologies is scattered among different sources, such as scientific journals, books, but also blogs, white papers and open online publishing platforms. The information available on informal sources was indeed very important to understand the perspective of real data science projects. In order to select articles for inclusion, a set of selection criteria was set:

\begin{itemize} \itemsep0.5em
	    
	\item Source and databases used for search:  \hyperlink{https://apps.webofknowledge.com}{Web of Science}, \hyperlink{https://www.mendeley.com/research-papers/}{Mendeley}, \hyperlink{https://scholar.google.es/}{Google Scholar}, \hyperlink{https://www.google.com}{Google}
	          
	\item Time period: from 2010 until 2020 (exception on CRISP-DM, from 1995)
	          
	\item Language: English
	          
	\item Document type: journal paper, conference paper, white paper. 
	      Both academic and practitioner papers were used. Practitioner articles were also included as they could provide an industry-level perspective on the data science \\ methodologies. 
	          
	\item Content: all papers selected had to do with data science project methodologies either directly (title, or text), or indirectly (inferred by the content). All the included papers involved findings related to at least one of the three categories of methodologies analyzed in this review (i.e. team, project and data \& information dimensions of data science project management). 
	          
\end{itemize}

Ultimately, a total of 19 studies (from 1996 to 2019) were selected for examination. Each study contained a lot of information, and thus, it was decided that the best way to compare studies was through the creation of a comparative table. This first table attempted to separate the key elements of the studies into four points:
\begin{enumerate}
	\item Paper details (Author / Journal / Year) 
	\item Main ideas and keywords
	\item Perspective (team, project, data \& information)
	\item Findings
\end{enumerate}

Afterwards, a second table was created to analyze how each methodology met the demands of the identified challenges. This table has been partitioned into three separate Tables [\ref{tab:project_management},\ref{tab:team_management},\ref{tab:data_information_management}], one for each category of challenges. The first column takes into account the challenges identified on  \cref{theoretical_framework} for team management, project management and data \& information management. In order to deal these challenges, methodologies design and implement different processes. Examples of such processes have been included on the second column. Based on the proposal of each methodology, a score is assigned to each criteria. These scores attempt to capture the effort put by the authors to solve each challenge. Taking as a reference \cite{brown2010likert}, the following punctuation system has been used:

\begin{itemize} \itemsep0em
	\item 0: criteria not fulfilled at all
	\item 1: criteria hardly fulfilled
	\item 2: criteria partially fulfilled
	\item 3: criteria fulfilled to a great extend
\end{itemize}

\begin{figure*}[hbt!]
	\centering
	\includegraphics[width=1\linewidth]{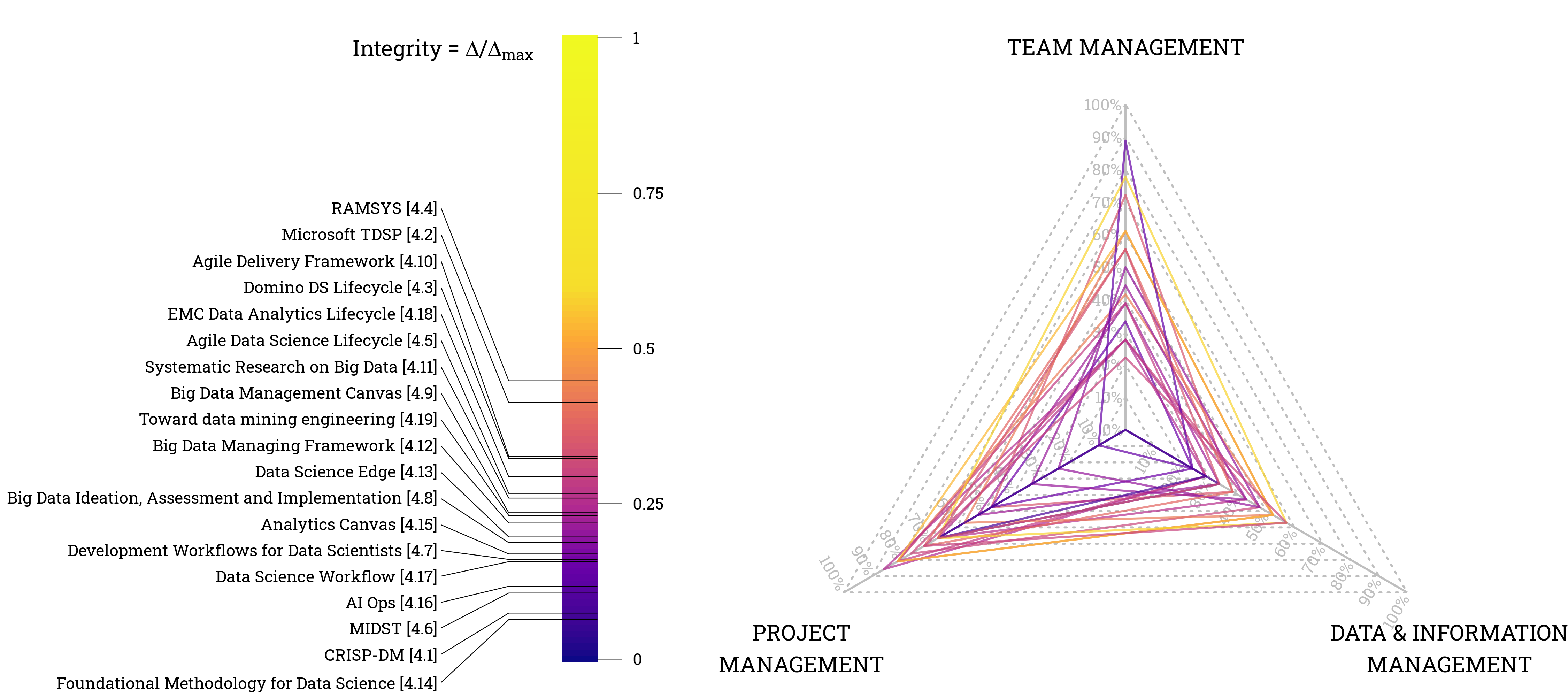}
	\caption{Quantitative summary of the reviewed methodologies: a) integrity value is represented on the bar plot and b) each category's scores are illustrated on the triangular plot, with the line color representing the integrity}
	\label{fig:joined}
\end{figure*}

The quantitative information included on the Tables [\ref{tab:project_management},\ref{tab:team_management},\ref{tab:data_information_management}] has also been illustrated in a triangular plot (\cref{fig:joined} right). Each axis represents a category of challenges - team, project, data \& information -  and the value in percentage (\%) reflects how well a methodology addresses those challenges. More specifically, the percentage value is calculated as the ratio between the sum of the scores and the maximum possible sum in each category. For example, the project management category has 7 challenges, so the maximum possible score is $7\cdot3 = 21$. A methodology that obtains 14 points will have a percentage score of $14/21=66\%$.

The percentage scores included in the triangular plot were also exploited to estimate the integrity of methodologies, which measures how well a given methodology covers all three categories of challenges. \cref{fig:joined} (left) illustrates the integrity scores for the reviewed methodologies. The integrity is calculated as the ratio between the area of the triangle and the area of the triangle with perfect scores $100\%$ on the three categories, which clearly has the maximum area ($3\sqrt{3}/4$). Calculating the area of the triangle using the three percentage scores is not straightforward, as it is explained in the \ref{appendixA}. 

\begin{table*}[hbt!]
    \large
    \centering
    \caption{Methodology Scores on Project Management}
    \resizebox{1\textwidth}{!}{
		\begin{tabular}{p{6.5cm}p{6cm}D{1.6em}D{1.6em}D{1.6em}D{1.6em}D{1.6em}D{1.6em}D{1.6em}D{1.6em}D{1.6em}D{1.6em}D{1.6em}D{1.6em}D{1.6em}D{1.6em}D{1.6em}D{1.6em}D{1.6em}D{1.6em}D{1.6em}}
            \toprule
            To address the challenges of...   & Methodology provides...  & {[}\ref{sec:crispdm}] & {[}\ref{sec:tdsp}] & {[}\ref{sec:domino}] & {[}\ref{sec:ramsys}] & {[}\ref{sec:agile_ds}] & {[}\ref{sec:midst}] & {[}\ref{sec:github}] & {[}\ref{sec:bigdata_ideation}] & {[}\ref{sec:bigdata_canvas}] & {[}\ref{sec:agile_delivery}] & {[}\ref{sec:systematic}] & {[}\ref{sec:bigdata_managing}] & {[}\ref{sec:dsedge}] & {[}\ref{sec:fmds}] & {[}\ref{sec:analytics_canvas}] & {[}\ref{sec:aiops}] & {[}\ref{sec:dsworkflow}] & {[}\ref{sec:emc}] & {[}\ref{sec:dm_engineering}] \\ \midrule
            Low level of process maturity & data science lifecycle: high level tasks / guideline / project management scheme & 3 & 3 & 3 & 3 & 3 & 0 & 3 & 3 & 2 & 3 & 3 & 3 & 3 & 3 & 2 & 2 & 3 & 3 & 3 \\[6ex]
			Uncertain business objectives & schematic to prevent making the important questions too late in time & 2 & 2 & 2 & 2 & 1 & 0 & 1 & 3 & 3 & 2 & 3 & 3 & 1 & 1 & 3 & 2 & 0 & 3 & 2 \\[6ex]
			Set adequate expectations & gives importance to the business or industry understanding phase & 3 & 3 & 2 & 3 & 2 & 0 & 2 & 3 & 3 & 2 & 2 & 3 & 2 & 3 & 2 & 2 & 1 & 2 & 3 \\[3.5ex]
			Hard to establish realistic project timelines & processes to control and monitor how long specific steps will take & 0 & 1 & 0 & 0 & 3 & 2 & 0 & 0 & 0 & 3 & 0 & 2 & 2 & 0 & 0 & 0 & 0 & 0 & 0 \\[3.5ex]
			Biased emphasis on technical issues & involved with aligning business requirements and data science goals & 2 & 2 & 3 & 2 & 0 & 0 & 1 & 3 & 3 & 3 & 2 & 0 & 2 & 1 & 2 & 2 & 1 & 2 & 2 \\[6ex]
			Delivering the "wrong thing" & makes sure that the outcome of the project is what the client/researcher is expecting & 2 & 3 & 1 & 2 & 1 & 0 & 0 & 3 & 3 & 2 & 2 & 3 & 2 & 3 & 0 & 2 & 0 & 2 & 2 \\[6ex]
			Results not used by business & methods for results evaluation, integration of the service or product in the client environment and provides necessary training & 2 & 3 & 1 & 2 & 0 & 0 & 0 & 3 & 2 & 2 & 2 & 3 & 2 & 3 & 2 & 0 & 0 & 3 & 3 \\ \midrule
			& \multicolumn{1}{r}{total}  & 14 & 17 & 12 & 14 & 10 & 2 & 7 & 18 & 16 & 17 & 14 & 17 & 14 & 14 & 11 & 10 & 5 & 15 & 15 \\
			& \multicolumn{1}{r}{\% over perfect (21)} & 67 & 81 & 57 & 67 & 48 & 10 & 33 & 86 & 76 & 81 & 67 & 81 & 67 & 67 & 52 & 48 & 24 & 71 & 71 \\ \bottomrule
		\end{tabular}
		}
	\label{tab:project_management}
\end{table*}
\begin{table*}[hbt!]
    \large
    \centering
    \caption{Methodology Scores on Team Management}
    \resizebox{1\textwidth}{!}{
		\begin{tabular}{p{6.5cm}p{6cm}D{1.6em}D{1.6em}D{1.6em}D{1.6em}D{1.6em}D{1.6em}D{1.6em}D{1.6em}D{1.6em}D{1.6em}D{1.6em}D{1.6em}D{1.6em}D{1.6em}D{1.6em}D{1.6em}D{1.6em}D{1.6em}D{1.6em}}
            \toprule
            To address the challenges of...   & Methodology provides...  & {[}\ref{sec:crispdm}] & {[}\ref{sec:tdsp}] & {[}\ref{sec:domino}] & {[}\ref{sec:ramsys}] & {[}\ref{sec:agile_ds}] & {[}\ref{sec:midst}] & {[}\ref{sec:github}] & {[}\ref{sec:bigdata_ideation}] & {[}\ref{sec:bigdata_canvas}] & {[}\ref{sec:agile_delivery}] & {[}\ref{sec:systematic}] & {[}\ref{sec:bigdata_managing}] & {[}\ref{sec:dsedge}] & {[}\ref{sec:fmds}] & {[}\ref{sec:analytics_canvas}] & {[}\ref{sec:aiops}] & {[}\ref{sec:dsworkflow}] & {[}\ref{sec:emc}] & {[}\ref{sec:dm_engineering}] \\ \midrule
            Poor coordination & establishes roles/responsibilities in a data science project & 0 & 3 & 3 & 3 & 3 & 3 & 3 & 2 & 1 & 0 & 2 & 0 & 1 & 0 & 2 & 2 & 1 & 3 & 0 \\[3.5ex]
			Collaboration issues across teams & defines clear tasks and problems for every person, and how to collaborate & 0 & 3 & 3 & 3 & 3 & 3 & 1 & 0 & 0 & 2 & 2 & 0 & 0 & 0 & 0 & 0 & 3 & 3 & 0 \\[6ex]
			Lack of transparent communication & describes how teams should work to communicate, coordinate and collaborate effectively & 0 & 2 & 0 & 2 & 2 & 3 & 3 & 0 & 0 & 1 & 2 & 0 & 2 & 0 & 0 & 0 & 3 & 0 & 0 \\[6ex]
			Inefficient governance models & describes how to coordinate with IT and approaches the deployment stage & 0 & 0 & 2 & 3 & 2 & 3 & 0 & 1 & 3 & 2 & 1 & 3 & 1 & 0 & 1 & 3 & 0 & 2 & 3 \\[6ex]
			Rely not only on leading data scientist & shares responsibilities, promotes training of employees & 0 & 1 & 2 & 1 & 2 & 2 & 0 & 2 & 0 & 1 & 2 & 1 & 1 & 0 & 2 & 0 & 0 & 2 & 0 \\[3.5ex]
			Build multidisciplinary teams & promotes teamwork of multidisciplinary profiles & 0 & 2 & 1 & 2 & 1 & 2 & 1 & 0 & 1 & 1 & 1 & 3 & 0 & 0 & 2 & 1 & 2 & 0 & 1 \\ \midrule
			&  \multicolumn{1}{r}{total} & 0 & 11 & 11 & 14 & 13 & 16 & 8 & 5 & 5 & 5 & 10 & 7 & 5 & 0 & 7 & 6 & 9 & 10 & 4 \\
			&  \multicolumn{1}{r}{\% over perfect (18)}  & 0 & 61 & 61 & 78 & 72 & 89 & 44 & 28 & 28 & 42 & 56 & 39 & 28 & 0 & 39 & 33 & 50 & 56 & 22  \\ \bottomrule
        \end{tabular}
	}
	\label{tab:team_management}
\end{table*}
\begin{table*}[hbt!]
    \large
    \centering
    \caption{Methodology Scores on Data \& Information Management}
    \resizebox{1\textwidth}{!}{
		\begin{tabular}{p{6.5cm}p{6cm}D{1.6em}D{1.6em}D{1.6em}D{1.6em}D{1.6em}D{1.6em}D{1.6em}D{1.6em}D{1.6em}D{1.6em}D{1.6em}D{1.6em}D{1.6em}D{1.6em}D{1.6em}D{1.6em}D{1.6em}D{1.6em}D{1.6em}}
            \toprule
            To address the challenges of... & Methodology provides... & {[}\ref{sec:crispdm}] & {[}\ref{sec:tdsp}] & {[}\ref{sec:domino}] & {[}\ref{sec:ramsys}] & {[}\ref{sec:agile_ds}] & {[}\ref{sec:midst}] & {[}\ref{sec:github}] & {[}\ref{sec:bigdata_ideation}] & {[}\ref{sec:bigdata_canvas}] & {[}\ref{sec:agile_delivery}] & {[}\ref{sec:systematic}] & {[}\ref{sec:bigdata_managing}] & {[}\ref{sec:dsedge}] & {[}\ref{sec:fmds}] & {[}\ref{sec:analytics_canvas}] & {[}\ref{sec:aiops}] & {[}\ref{sec:dsworkflow}] & {[}\ref{sec:emc}] & {[}\ref{sec:dm_engineering}] \\ \midrule
			Lack of reproducibility & offers setup for assuring reproducibility and traceability & 0 & 3 & 3 & 1 & 2 & 2 & 3 & 0 & 3 & 0 & 0 & 0 & 0 & 0 & 0 & 0 & 3 & 0 & 2 \\[4ex]
			Retaining and accumulation of knowledge & generation and accumulation of knowledge: data, models, experiments, project insights, best practices and pitfalls & 2 & 2 & 3 & 3 & 2 & 3 & 3 & 0 & 3 & 1 & 0 & 1 & 0 & 0 & 1 & 0 & 3 & 0 & 2 \\[9ex]
			Low data quality for Machine Learning algorithms & takes into account the limitations of Machine Learning techniques & 1 & 0 & 0 & 2 & 0 & 0 & 0 & 0 & 0 & 0 & 0 & 0 & 2 & 1 & 0 & 1 & 2 & 1 & 2 \\[3.5ex]
			Lack of quality assurance checks & tests to check data limitations in quality and potential use & 2 & 3 & 0 & 2 & 1 & 0 & 1 & 0 & 0 & 3 & 2 & 0 & 2 & 2 & 0 & 2 & 2 & 3 & 2 \\[3.5ex]
			No validation data & robust validation of the proposed solution \& makes hypothesis & 2 & 2 & 3 & 2 & 2 & 0 & 0 & 3 & 1 & 2 & 3 & 2 & 1 & 1 & 2 & 0 & 0 & 0 & 2 \\[3.5ex]
			Data security and privacy & concerned about data security and privacy & 0 & 0 & 0 & 2 & 0 & 0 & 2 & 0 & 0 & 2 & 0 & 0 & 2 & 0 & 2 & 0 & 0 & 2 & 0 \\[3.5ex]
			Investment in IT infrastructure & preallocates resources for investment on IT resources & 0 & 1 & 2 & 0 & 1 & 0 & 0 & 3 & 3 & 3 & 2 & 3 & 2 & 2 & 2 & 2 & 0 & 2 & 2 \\ \midrule
			& \multicolumn{1}{r}{total} & 7 & 11 & 11 & 12 & 8 & 5 & 9 & 6 & 10 & 11 & 7 & 6 & 9 & 6 & 7 & 5 & 10 & 8 & 12 \\
			& \multicolumn{1}{r}{\% over perfect (21)} & 33 & 52 & 52 & 57 & 38 & 24 & 43 & 29 & 48 & 52 & 33 & 29 & 43 & 29 & 33 & 24 & 48 & 38 & 57 \\ \bottomrule
		\end{tabular}
	}
	\label{tab:data_information_management}
	\resizebox{1\textwidth}{!}{
		\begin{tabular}{@{}l@{}}
			\\Score Punctuation System: criteria not fulfilled at all (0), hardly fulfilled (1), partially fulfilled (2), fulfilled to a great extend (3).
			\\
			\\Legend: 4.1 CRISP-DM; 4.2 Microsoft TDSP; 4.3 Domino DS Lifecycle; 4.4 RAMSYS; 4.5 Agile Data Science Lifecycle; 4.6 MIDST; 4.7 Development Workflows for Data \\ Scientists; 4.8 Big Data Ideation, Assessment and Implementation; 4.9 Big Data Management Canvas; 4.10 Agile Delivery Framework; 4.11 Systematic Research on Big Data;\\ 4.12 Big Data Managing Framework; 4.13 Data Science Edge; 4.14 Foundational Methodology for Data Science; 4.15 Analytics Canvas; 4.16 AI Ops; 4.17 Data Science \\ Workflow; 4.18 EMC Data Analytics Lifecycle; 4.19 Toward data mining engineering
		\end{tabular}
	}
	
\end{table*}

Therefore, Tables [\ref{tab:project_management},\ref{tab:team_management},\ref{tab:data_information_management}] quantitatively evaluate the qualitative characteristics of each methodology and its suitability to take on the challenges identified on \cref{theoretical_framework} for the management of teams, projects and data \& information.

\section{Results: Critical Review of Data Science Methodologies}    \label{results}

Hereunder we review a collection of methodologies that propose guidelines for managing and executing data science projects. Each methodology is described in detail in order to analyze how well it meets the demands of the presented challenges. This includes a description of structure, features, principles, artifacts, and recommendations. Each methodology is concluded with a summary of its main strengths and drawbacks. We suggest the reader to check the scores given to each methodology after reading the corresponding critical review. 

\subsection{CRISP-DM} \label{sec:crispdm}

\noindent
Cross-Industry Standard Process for Data Mining (CRISP-DM) \cite{shearer_crispdm} is an open standard process model developed by SPSS and Teradata in 1996 that describes common approaches used by data mining experts. It presents a structured, well defined and extremely documented process iterative process. CRISP-DM breaks down the lifecycle of a data mining project into six phases: business understanding, data understanding, data preparation, modeling, evaluation and deployment. 

The business understanding phase focuses on the project objectives and the requirements from a business perspective, and converts this information into a data science problem. Thus it tries to align business and data science objectives, setting adequate expectations and focusing of delivering what the business is expecting. The outcome of this stage is usually a fixed project plan, which hardly takes into account the difficulty to establish realistic project timelines due to the exploratory nature of data science projects and their intrinsic uncertainty.

The rest of the stages (data preparation, modeling, evaluation and deployment) are quite straight-forward, and the reader is surely familiar with them. An strict adherence to the CRISP-DM process forces the project manager to document the project and the decisions made along the way, thus retaining most of the generated knowledge. Apart from that, the data preparation phase involves data quality assessment tests, while the evaluation phase gives a noticeable importance to the validation and evaluation of the project results.

Nevertheless, one of the main shortcomings of CRISP-DM is that it does not explain how teams should organize to carry out the defined processes and does not address any of the above mentioned team management issues. In this sense, in words of \cite{grady_kdd}, CRISP-DM needs a better integration with management processes, demands to align with software and agile development methodologies, and instead of simple checklists, it also needs method guidance for individual activities within stages.

Overall, CRISP-DM provides a coherent framework for guidelines and experience documentation. The data science lifecycle presented by CRISP-DM is commonly used as a reference by other methodologies, that replicate it with different variations. 
However, CRISP-DM was conceived in 1996, and therefore is not updated to the current paradigm and the latest developments in data science technologies, especially in regard of big data advancements. In words of industry veteran Gregory Piatetsky of KDNuggets: ``CRISP-DM remains the most popular methodology for analytics, data mining, and data science projects, with 43\% share in latest KDnuggets survey \cite{piatetsky_crispdm}, but a replacement for unmaintained CRISP-DM is long overdue''. 

\begin{itemize}
\item[$\color{green}\bigtriangleup$] Coherent and well documented iterative process
\item[$\color{red}\bigtriangledown$] Does not explain how teams should organize and does not address any team management issues
\end{itemize}

\subsection{Microsoft TDSP} \label{sec:tdsp}

\noindent
Microsoft Team Data Science Process (TDSP) by Microsoft \cite{microsoft_tdsp} is an ``agile, iterative data science methodology that helps improving team collaboration and learning''. It is very well documented and it provides several tools and utilities that facilitate its use. Unfortunately, TDSP is very dependent on Microsoft services and policies, and this complicates a broader use. In fact, we believe that any methodology should be independent of any tools or technologies. As it was defined on \cite{ellis_frameworks_methodologies}, a methodology is a general approach that guides the techniques and activities within a specific domain with a defined set of rules, methods and processes and does not rely on certain technologies or tools.
With independence of the Microsoft tools in which TSDP relies, this methodology provides some interesting processes, both on project, team and data \& information management, which have been summarized.

The key components of TDSP are: 
1) data science lifecycle definition
2) a standardized project structure
3) infrastructure and resources and
4) tools and utilities for project execution.

TDSP’s project lifecycle is like CRISP-DM and includes five iterative stages: a) Business Understanding b) Data Acquisition and Understanding c) Modeling d) Deployment and e) Customer Acceptance.
It is indeed a iterative and cyclic process: the output of the ``Data Acquisition and Understanding'' phase can feed back to the ``Business Understanding'' phase, for example.

At the business understanding stage, TDSP is concerned about defining SMART (Specific, Measurable, Achievable, Relevant, Time-bound) objectives and identifying the data sources. In this sense, one of the most interesting artifacts is the charter document: this standard template is a living document that keeps updating throughout the project as new discoveries are made and as business requirements change as well. This artifact helps documenting the project discovery process, and also promotes transparency and communication, as long as stakeholders are involved in it. Besides, iterating upon the charter document facilitates the generation and accumulation of knowledge and valuable information for future projects. Along with other auxiliary artifacts, this document can help tracing back on the history of the project and reproducing different experiments. Regarding reproducibility, for each tested model, TDSP provides a model report: a standard, template-based report with details on each experiment.

In relation to the quality assessment, a data quality report is prepared during the data acquisition and understanding phase. This report includes data summaries, the relationships between each attribute and target, variable ranking, and more. To ease this task, TDSP provides an automated utility, called IDEAR, to help visualize the data and prepare data summary reports. In the final phase the customer validates whether the system meets their business needs and whether it answers the questions with acceptable accuracy. 

TDSP addresses the weakness of CRISP-DM’s lack of team definition by defining four distinct roles (solution architect, project manager, data scientist, and project lead) and their responsibilities during each phase of the project lifecycle. These roles are very well defined from a project management perspective and the team works under Agile methodologies, which improve the collaboration and the coordination. Their responsibilities regarding the project creation, execution and development are clear.

\begin{itemize}
\item[$\color{green}\bigtriangleup$] Integral methodology: provides processes both on project, team and data \& information management
\item[$\color{red}\bigtriangledown$] Excessive dependence on Microsoft tools and technologies
\end{itemize}

\begin{figure*}[bt!]
	\centering
	\begin{minipage}{0.24\textwidth}
		\includegraphics[width=\linewidth]{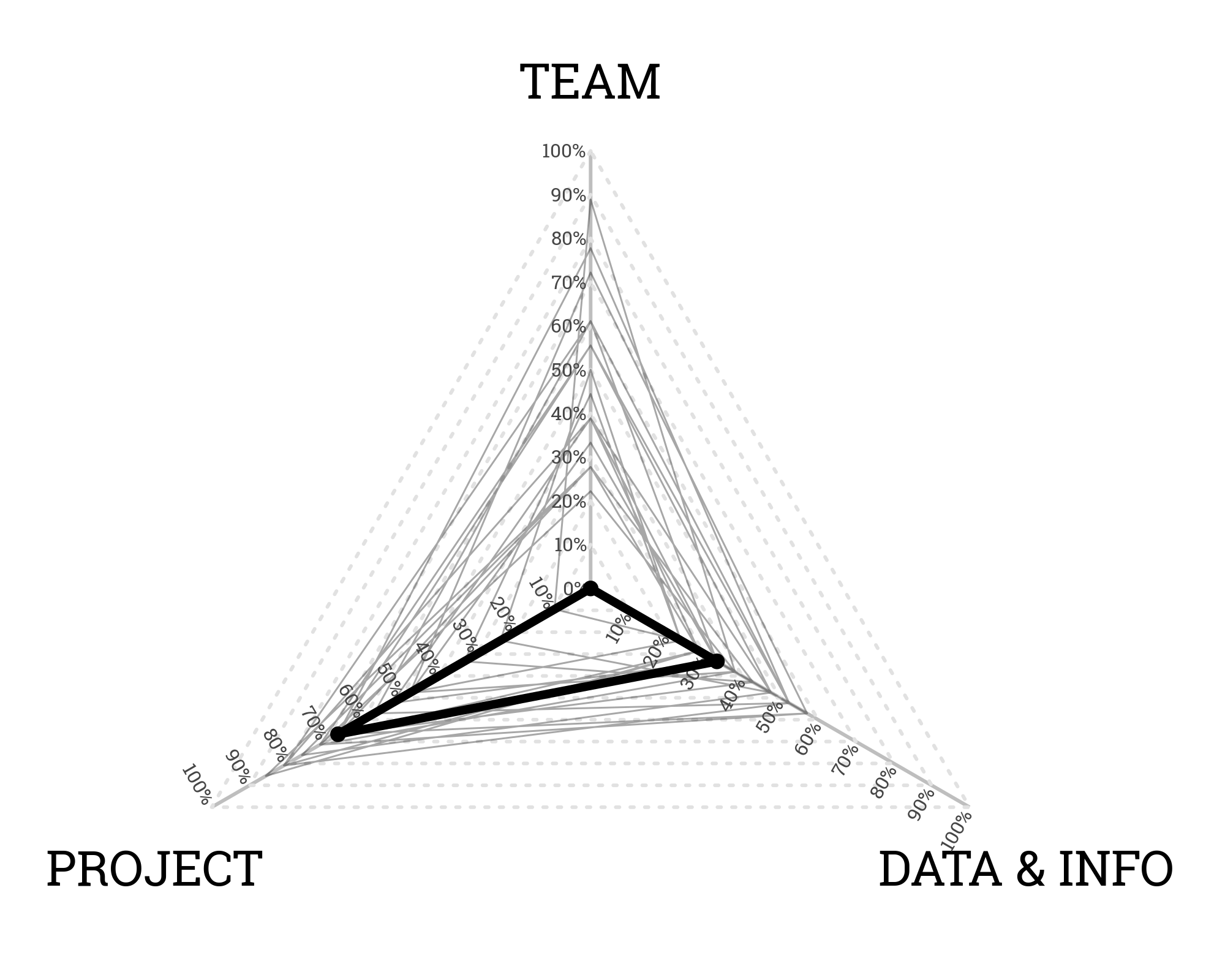}
		\caption{CRISP-DM} \label{crispdm}
	\end{minipage}\hfill
	\begin{minipage}{0.24\textwidth}
    		\includegraphics[width=\linewidth]{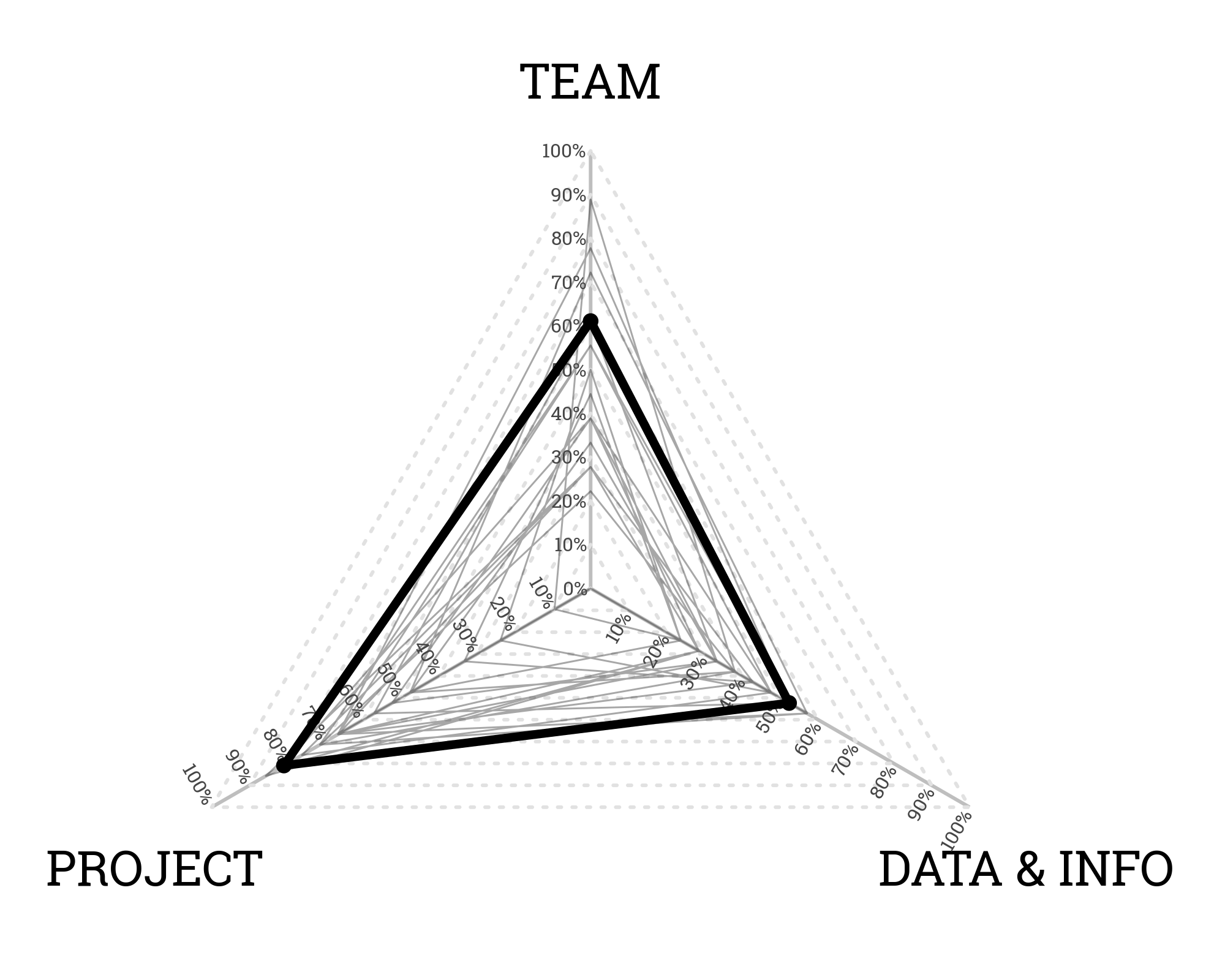}
    		\caption{Microsoft TDSP} \label{tdsp}
  	\end{minipage}\hfill
  	\begin{minipage}{0.24\textwidth}
    		\includegraphics[width=\linewidth]{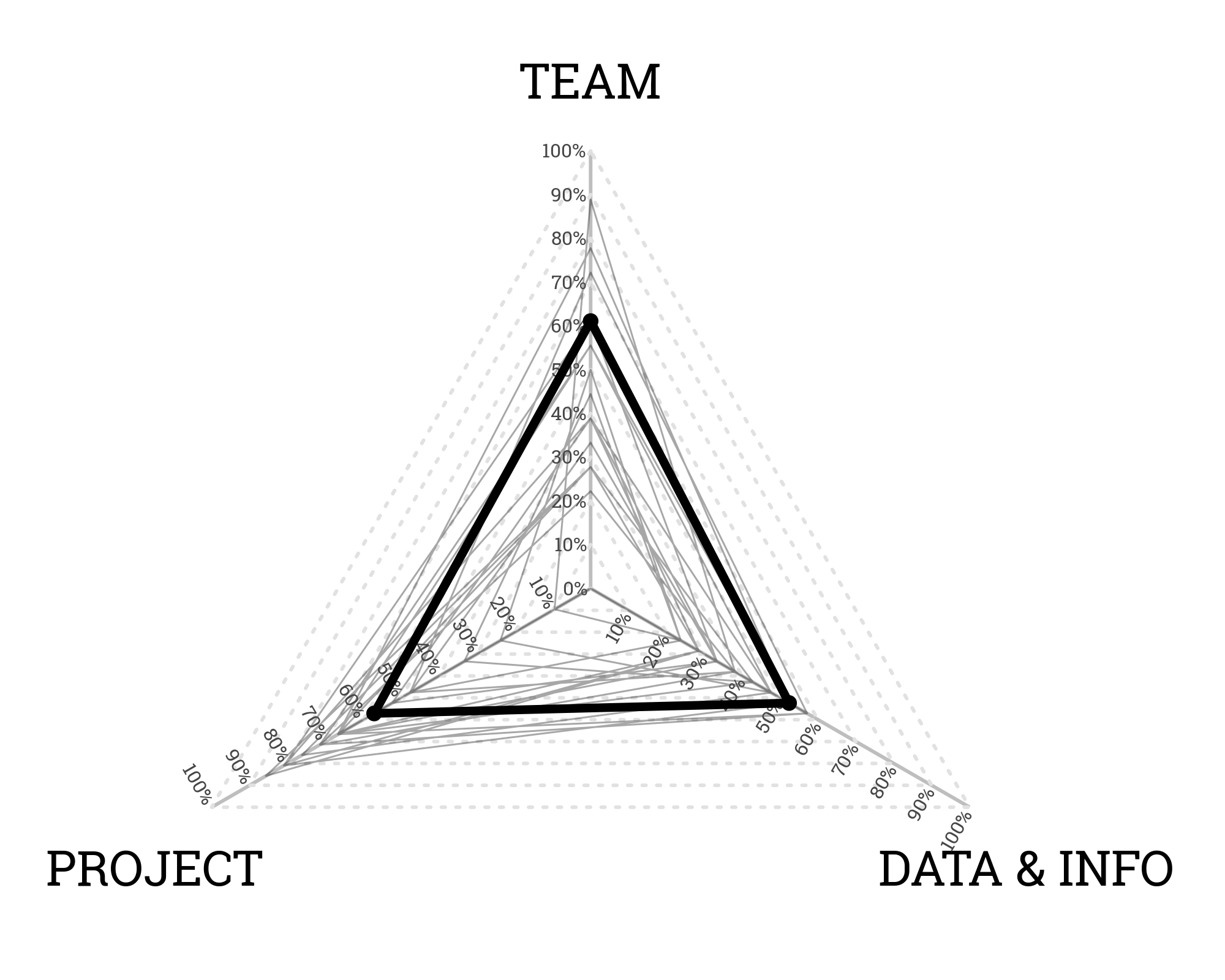}
    		\caption{Domino DS Lifecycle} \label{domino}
  	\end{minipage}\hfill
  	\begin{minipage}{0.24\textwidth}
    		\includegraphics[width=\linewidth]{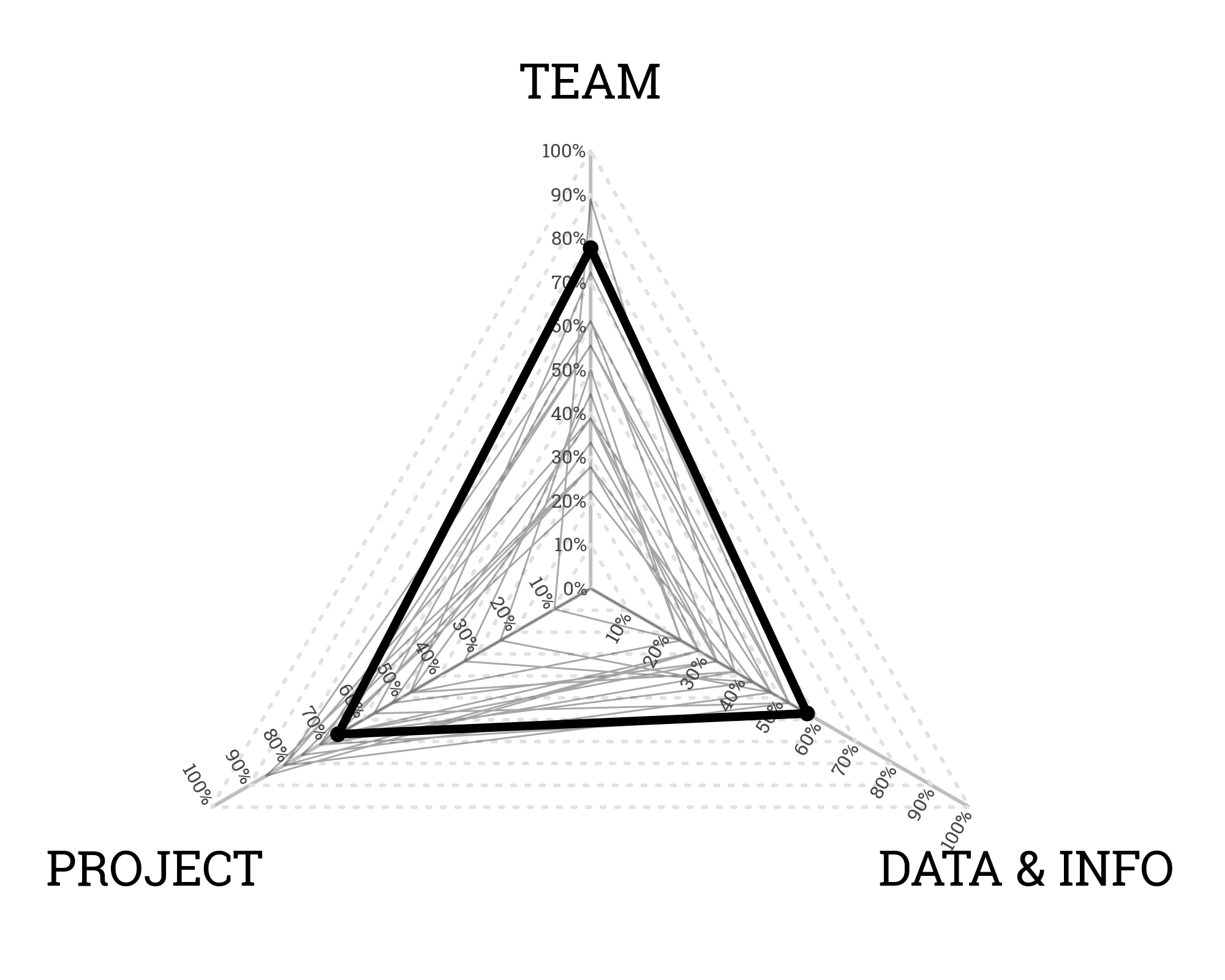}
		\caption{RAMSYS} \label{ramsys}
  	\end{minipage}
\end{figure*}

\subsection{Domino DS Lifecycle} \label{sec:domino}

\noindent
Domino Data Lab introduced its data science project lifecycle in a 2017 white-paper \cite{domino_data_lab}. Inspired by CRISP-DM, agile, and a series of client observations, it takes ``a holistic approach to the entire project lifecycle from ideation to delivery and monitoring.''
The methodology is founded on three guiding principles: 
a) “Expect and embrace iteration” but “prevent iterations from meaningfully delaying projects, or distracting them from the goal at hand”.
b) “Enable compounding collaboration” by creating components that are re-usable in other projects.
c) “Anticipate auditability needs” and “preserve all relevant artifacts associated with the development and deployment of a model”.

The proposed data science lifecycle is composed of the following phases: a) ideation b) data acquisition and exploration c) research and development d) validation e) delivery f) monitoring

The ideation mirrors the business understanding phase from CRISP-DM. In this phase business objectives are identified, success criteria are outlined and if it is the case, the initial ROI analysis is performed. This phase also integrates common agile practices including developing a stakeholder-driven backlog and creating deliverable mock-ups. These preliminary business analysis activities can dramatically reduce the project risk by driving alignment across stakeholders. 

The data acquisition and exploration phase incorporates many elements from the data understanding and data preparation phases of CRISP-DM. The research and development phase is the ``perceived heart'' of the data science process. It iterates through hypothesis generations, experimentation, and insight delivery. DominoLab recommends starting with simple models, setting a cadence for insight deliveries, tracking business KPIs, and establishing standard hardware and software configurations while being flexible to experiment.

The validation stage highlights the importance of ensuring reproducibility of results, automated validation checks, and the preservation of the null results. In the delivery stage, Domino recommends to preserve links between deliverable artifacts, flag dependencies, and develop a monitoring and training plan. Considering models’ non deterministic nature, Domino suggests monitoring techniques that go beyond standard software monitoring practices: for example, using control groups in production models to keep track of model performance and value creation to the company.

Overall, Domino’s model does not describe every single step but is more informative to lead a team towards better performance. It effectively integrates data science, software engineering, and agile approaches. Moreover, it leverages agile strategies, such as fast iterative deliveries, close stakeholder management, and a product backlog.
In words of \cite{hotz_dominolab} Domino’s lifecycle should not be viewed as mutually exclusive with CRISP-DM or Microsoft’s TDSP; rather its “best practices” approach with “a la carte” elements could augment these or other methodologies as opposed to replace them.

\begin{itemize}
\item[$\color{green}\bigtriangleup$] Holistic approach to the project lifecycle. Effectively integrates data science, software engineering, and agile approaches
\item[$\color{red}\bigtriangledown$] Informative methodology rather than prescriptive
\end{itemize}

\subsection{RAMSYS} \label{sec:ramsys}

\noindent
RAMSYS \cite{moyle_ramsys} by Steve Moyle is a methodology for supporting rapid remote collaborative data mining projects. It is intended for distributed teams and the principles that guide the design of the methodology are: light management, start any time, stop any time, problem solving freedom, knowledge sharing, and security.

This methodology follows and extends the CRISP-DM methodology, and it allows the data mining effort to be expended at very different locations communicating via a web-based tool. The aim of the methodology is to enable information and knowledge sharing, as well as the freedom to experiment with any problem solving technique.

RAMSYS defines three roles, named ``modellers'', ``data master'' and ``management committee''. The ``data master'' is responsible for maintaining the database and applying the necessary transformations. The ``management committee'' ensures that information flows within the network and that a good solution is provided. This committee is also responsible for managing the interface with the client and setting up the challenge relative to the data science project in hand: defining the success criteria, receiving and selecting submissions. In this way, ``modellers'' experiment, test the validity of each hypothesis and produce new knowledge. With that, they can suggest new data transformations to the ``data master''. One of the strengths of the RAMSYS model is the relative freedom given to modellers in the project to try their own approaches. Modellers benefit of central data management, but have to conform to the defined evaluation criteria. 

The ``information vault'' is an proposed artifact that contains the problem definition, the data, and the hypothesis about the data and the models, among other concepts. Moreover, RAMSYS proposes the use of the ``hypothesis investment account'', which contains all the necessary operations to refute or corroborate an hypothesis: the hypothesis statement, refutation/corroboration evidence, proof, hypothesis refinement and generalization. This artifact can be used to extract reusable knowledge in the form of lessons learned for future projects.

Therefore, RAMSYS intends to allow the collaborative work of remotely placed data scientists in a disciplined manner in what respects the flow of information while allowing the free flow of ideas for problem solving. Of course, this methodology can be applied to more common data science team that share location. The clear definition of the responsibilities for each role allows effective collaboration and the tools that support this methodology are well defined to be applied to real projects.

\begin{itemize}
\item[$\color{green}\bigtriangleup$] Considers distributed teams and enables information and knowledge sharing
\item[$\color{red}\bigtriangledown$] Lack of support for sharing datasets and models
\end{itemize}

\subsection{Agile Data Science Lifecycle} \label{sec:agile_ds}

\noindent
Agile Data Science Lifecycle \cite{jurney_agile_ds} by Russell Jurney presents a framework to perform data science combined with agile philosophy. This methodology asserts that the most effective and suitable way for data science to be valuable for organizations is through a web application, and therefore, under this point of view, doing data science evolves into building applications that describe the applied research process: rapid prototyping, exploratory data analysis, interactive visualization and applied machine learning. 

The book includes the ``Agile Data Science manifesto'', which attempts to apply agility to the practice of data science, and is based on several principles:
1) Iterate, iterate, iterate: the author emphasizes the iterative nature of the creating, testing and training learning algorithms. 2) Ship intermediate output: throughout cycles of iteration, any intermediate output is committed and shared with other team members. In this sense, by sharing work on a continual basis, the methodology promotes feedback and the creation of new ideas. 3) Prototype experiments over implementing tasks: it is uncertain if an experiment will lead to any valuable insight. 4) Integrate the tyrannical opinion of data in product management: it is important to listen to the data. 5) Climb up and down the data value pyramid: the data value pyramid provides a process path from initial data collection to discovering useful actions. The generation of value increases as the team climbs to higher layers of the pyramid. 6) Discover and pursue the critical path to a killer product. The critical path is the one that leads to something actionable that creates value. 7) Describe the process, not just the end state. Document the process to understand how the critical path was found.

Overall, Agile Data Science tries to align data science with the rest of the organization. Its main goal is to document and guide exploratory data analysis to discover and follow the critical path to a compelling product. The methodology also takes into account that products are built by teams of people, and hence it defines a broad spectrum of team roles, from customers to the DevOps engineers.

\begin{itemize}
\item[$\color{green}\bigtriangleup$] Rapid delivery of value from data to customer. \\
More realistic feedback: assess value in deliverables
\item[$\color{red}\bigtriangledown$] Agile is less straightforward, works better in dynamic environments and evolving requirements
\end{itemize}

\subsection{MIDST} \label{sec:midst}

\noindent
Kevin Crowston \cite{crowston_midst} presents a theoretical model of socio-technical affordances for stigmergic coordination. The goal of this methodology is to better support coordination in data science teams by transferring findings about coordination from free/libre open source software (FLOSS) development. For that purpose, the authors design and implement a system to support stigmergic coordination. 

A process is stigmergic if the work done by one agent provides a stimulus (stigma) that attracts other agents to continue the job. Thus, stigmergic coordination is a form of coordination that is based on signals from shared work. The organization of the collective action emerges from the interaction of the individuals and the evolving environment, rather than from a shared plan or direct interaction. 

As it is claimed on the article, the specific tools that would be useful to data science teams might be different than the features that are important for FLOSS dev teams, making it difficult to configure a data science collaboration environment using existing tools. That is why the authors propose an alternative approach based on stigmergic coordination and developed a web-based data science application to support it. In this regard, a theory of affordance to support stigmergic coordination is developed, which is based on the following principles: visibility of work, use of clear genres of work products and combinability of contributions.

The proposed framework is targeted to improve the collaboration between data scientists, and does not attempt to cover project management nor data management issues. It should be taken into consideration that the presented web-based application was further tested on real cases with students. While it is true that some of the reviewed methodologies are explained together with use cases, most lack empirical and field experimentation. That is why the effort put by this methodology in that direction is specially emphasized.

\begin{itemize}
\item[$\color{green}\bigtriangleup$] Improves collaboration between data scientists through stigmergic coordination.
\item[$\color{red}\bigtriangledown$] Exclusively focused on improving team coordination
\end{itemize}

\begin{figure*}[bt!]
	\centering
	\begin{minipage}{0.24\textwidth}
		\includegraphics[width=\linewidth]{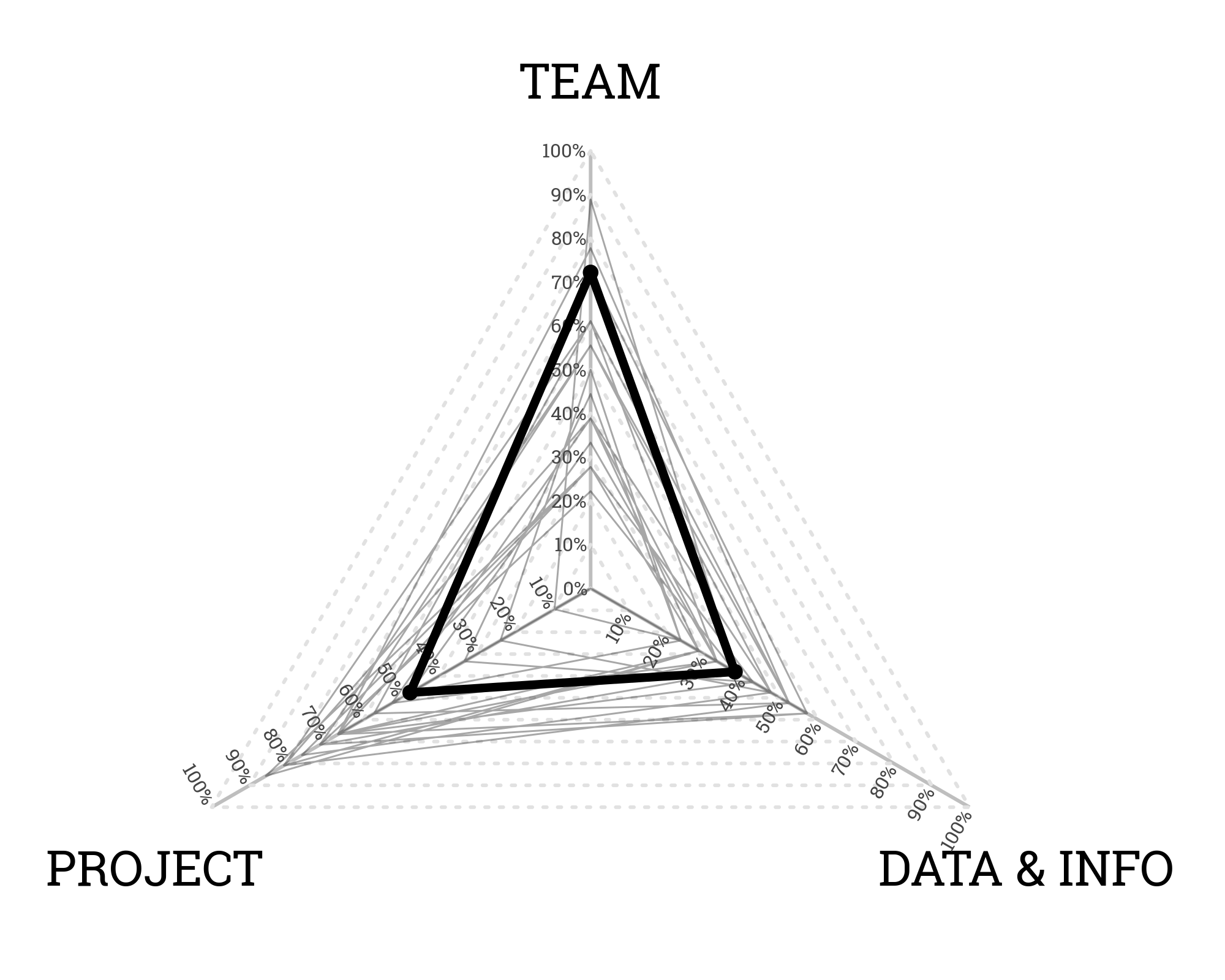}
    		\caption{Agile Data Science Lifecycle} \label{agile_ds}
	\end{minipage}\hfill
	\begin{minipage}{0.24\textwidth}
    		\includegraphics[width=\linewidth]{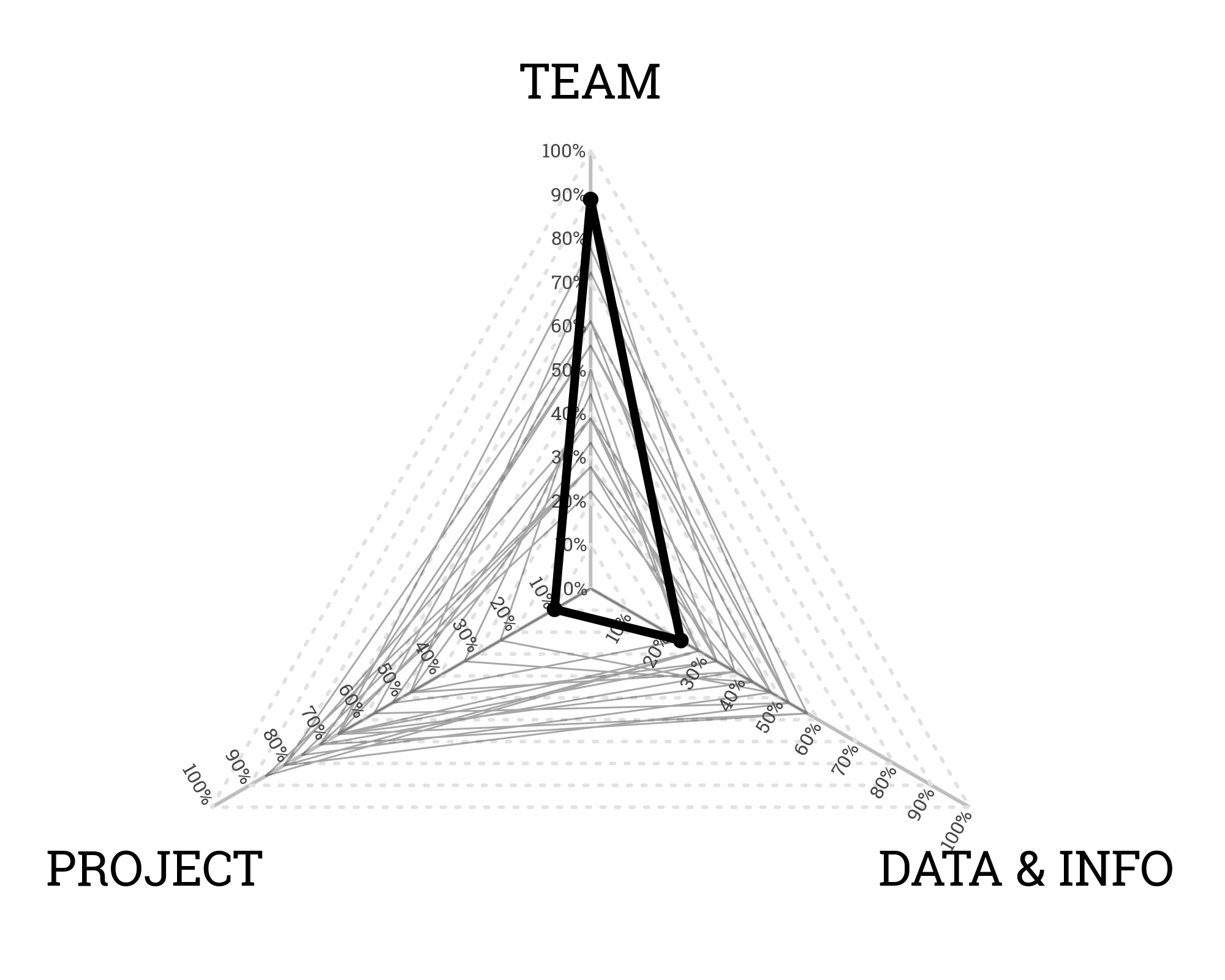}
    		\caption{MIDST} \label{midst}
  	\end{minipage}\hfill
  	\begin{minipage}{0.24\textwidth}
    		\includegraphics[width=\linewidth]{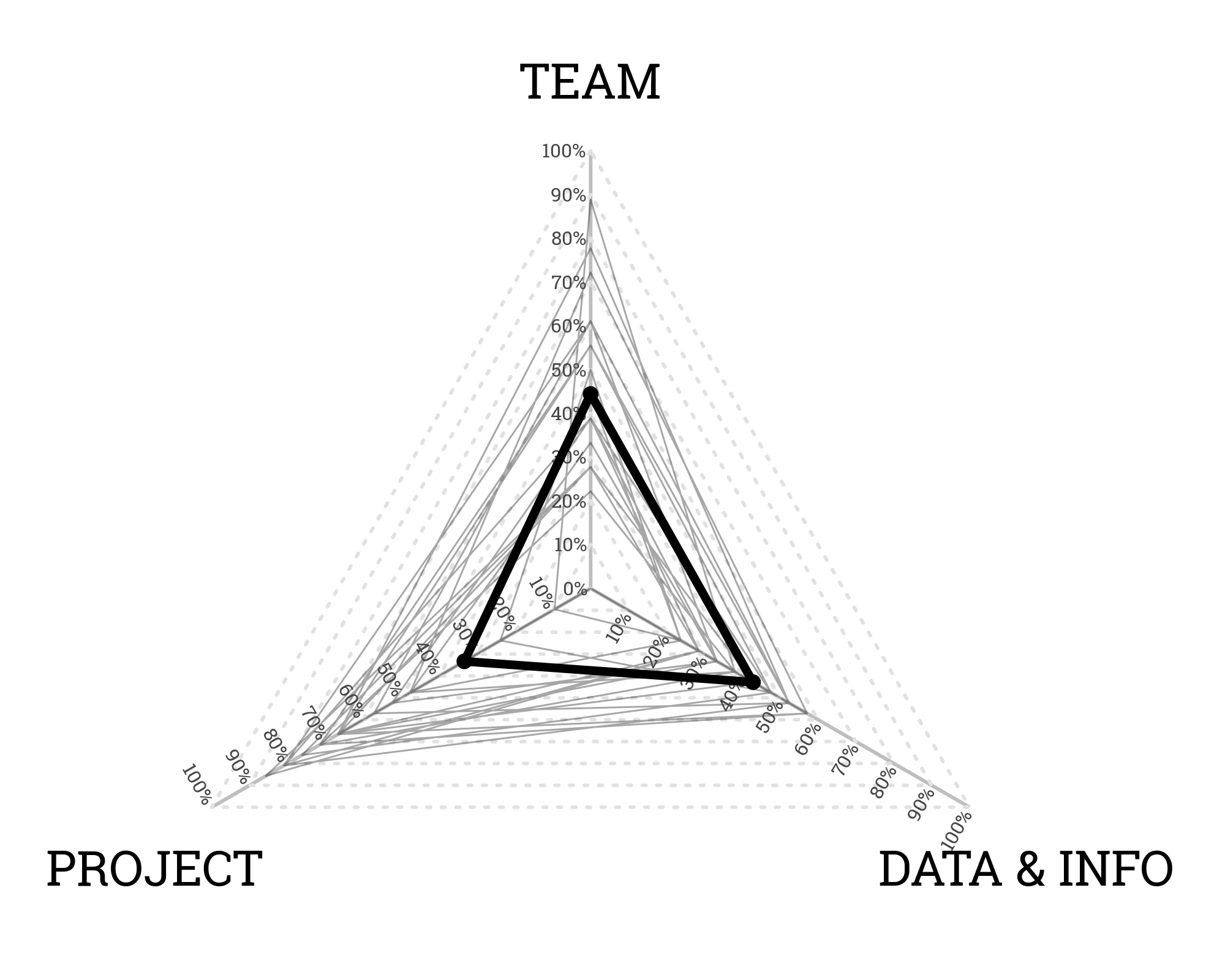}
		\caption{Development Workflows for Data Scientists} \label{github}
  	\end{minipage}\hfill
  	\begin{minipage}{0.24\textwidth}
    		\includegraphics[width=\linewidth]{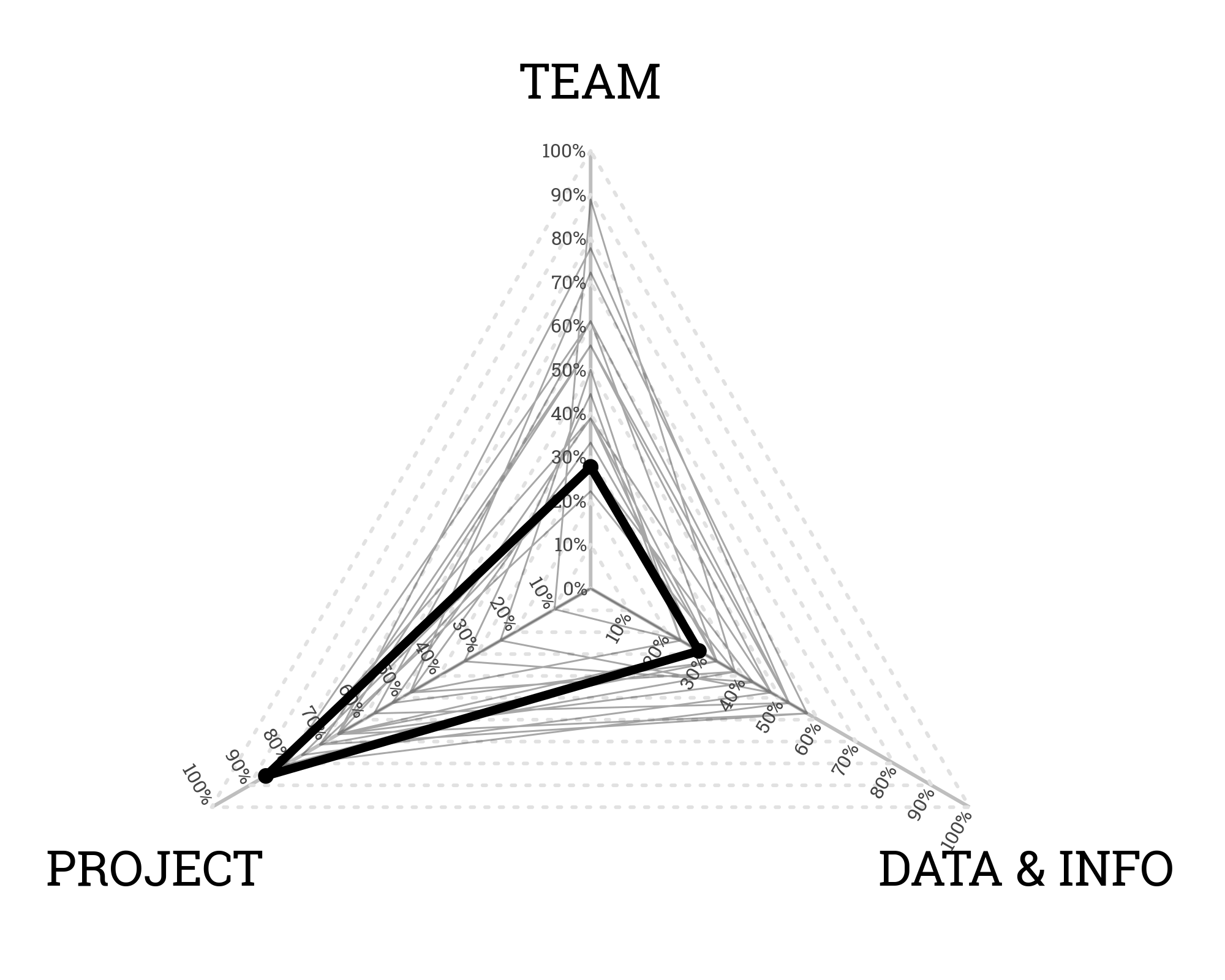}
    		\caption{Big Data Ideation, Assessment and Implementation} \label{bigdata_ideation}
  	\end{minipage}
\end{figure*}

\subsection{Development Workflows for Data Scientists} \label{sec:github}

\noindent
Development Workflows for Data Scientists \cite{byrne_development_workflows} by Github and O'Reilly Media gathers various best practices and workflows for data scientists. The document examines how several data-driven organizations are improving development workflows for data science. 

The proposed data science process follows an iterative structure: a) ask interesting question b) examine previous work c) get data d) explore the data e) model the data f) test g) document the code h) deploy to production i) communicate results

The process starts by asking an interesting question, which is described as one of the hardest tasks in data science. Understanding the business goals and the data limitations are required prior to asking interesting questions. In this regard, defining a suitable success measure for both the business and the data science team is also described as a challenge. The next step in the process is examining previous work. However, more often than not data science teams are coming across non-structured projects and scattered knowledge, which complicates understanding previous work. In this sense, they recommend the use of tools to make data science work more discoverable, such as the Airbnb's \hyperlink{https://github.com/airbnb/knowledge-repo/}{Knowledge Repo} \cite{airbnb_knowledge_repo}. 

From then on, data is obtained and explored. Regarding the data collection, the authors claim that is not enough to just gather relevant data, but to also understand how it was generated and deal with security, compliance and anonymization aspects. To help during the data exploration, it is recommended to have a good data directory structure. Tools such as \hyperlink{https://github.com/cookiecutter/cookiecutter}{cookiecutter} can take care of all the setup and boilerplate for data science projects.

With reference to the data modeling, it is suggested to create two teams: one for building models and a completely independent one to evaluate and validate the models. Doing so, data training leaks are prevented and also the success criteria is kept safe from the model creation process. The next step on the process, testing, is yet a debated area. Here is where data science and software development practices deviate: in data science there is a lot of trial and error, which can be incompatible with test-driven frameworks. Still, it is recommended to use testing technologies to improve interpretability, accuracy, and the user experience.

Having a documentation phase is without a doubt an innovative yet logical point for data science methodologies. In fact, documenting the working solution is not always enough, since it is equally valuable to know the pitfalls and the dead ends. It is recommended to create a codebook to record all steps taken, tools used, data sources, results and conclusions reached. This phase is directly related to the ``examining previous work'' phase, as having a well documented project will help subsequent projects. Then models are deployed to production. For this task, it is proposed to use an standard git-flow to version all changes to the models. Finally, the last step in the process is to communicate the results, which together with asking the initial question, can be the most problematic steps.

This methodology is built upon the recommendations from first-class data-driven organizations, and there lies its main strength. 
It includes novel phases in the data science workflow, such as ``previous work examination'', ``code documentation'' and ``results communication''. It also proposes a team structure and roles: data scientist, machine learning engineer, data engineer. Overall, it is concluded that ultimately a good workflow depends on the tasks, goals, and values of each team, but it recommends to produce results fast, reproduce, reuse and audit results, and enable collaboration and knowledge sharing.

\begin{itemize}
\item[$\color{green}\bigtriangleup$] Built upon recommendations and best-practices from first-class data-driven organizations
\item[$\color{red}\bigtriangledown$] Lacks an in-depth breakdown of each phase
\end{itemize}

\subsection{Big Data Ideation, Assessment and Implementation} \label{sec:bigdata_ideation}

\noindent
Big data ideation, assessment and implementation by Martin Vanauer \cite{vanauer_bigdata_organizations} is a methodology to guide big data idea generation, idea assessment and implementation management. It is based on big data 4V's (volume, variety, velocity, value and veracity), IT value theory, workgroup ideation processes and enterprise architecture management.

The authors claim that the introduction of big data resembles an innovative process, and therefore they follow a model of workgroup innovation to propose their methodology, which is structured in two phases: ideation and implementation. Ideation refers to the generation of solutions by applying big data to new situations. Implementation is related to the evaluation of the developed solutions and the subsequent realization. 

More specifically, for the ideation phase two perspectives are defined: either there are business requirements that can be better fulfilled by IT, which the authors called ``Business First'' (BF), or, the IT department opens up new business opportunities, so called ``Data First'' (DF). Under this perspective, there is no such thing as ``Technology First''. The authors claim that prioritizing technology is not feasible because technology should not be introduced as an end by itself, but in a way in which it provides value to the business.Therefore, the ideation phase either resembles the identification of business needs (BF), or the development of new business models based on an identification of available data (DF).

Each phase (ideation and implementation) is also structured into the transition and action sub-phases:
1) Ideation transition: comprises the mission definition and identification of objectives. Two artifacts are used to support this stage: the ``Modelling and Requirements Engineering'' for the BF and the ``Key Resources Assessment'' for the DF perspective.
2) Ideation action: here  creative solutions are identified by preparing a business use case (BF) or a value proposition (DF), assisted by a ``Business Model Canvas''.
3) Implementation transition: involves evaluating and selecting the most appropriate ideas. Here is where a financial and organizational feasibility study is carried out, by means of a cost/benefit analysis (BF) or a value proposition fit assessment (DF). The technical feasibility is also analyzed using EAM methods for impact assessment and rollout.
4) Implementation action: finally, the implementation roadmap is developed.

This proposal introduces a totally new perspective that has nothing to do with the rest of presented methodologies. It shows a completely different pipeline, that breaks away from the CRISP-DM-based data analytics lifecycles. Its main strength comes from the structure of ideation and implementation phases, but more importantly from the distinction of business-first vs data-first perspectives, which define the subsequent paths to be followed. However, it does not address team related challenges, nor data management issues.

\begin{itemize}
\item[$\color{green}\bigtriangleup$] Distinction of business-first vs data-first perspectives
\item[$\color{red}\bigtriangledown$] Overlooks team related and data management issues
\end{itemize}

\subsection{Big Data Management Canvas} \label{sec:bigdata_canvas}

\noindent
Big Data Management Canvas by Michael Kaufmann \cite{kaufmann_bigdata_canvas} is a reference model for big data management that operationalizes value creation from data by linking business targets with technical implementation. 

This research article proposes a big data management process that helps knowledge emerge and create value. Kaufmann asserts that creating value from big data relies on ``supporting decisions with new knowledge extracted from data analysis''. It provides a frame of reference for data science to orient the engineering of information systems for big data processing, toward the generation of knowledge and value. The author incorporates the value term on the big data paradigm, in this case defined by the 5V's: volume, velocity, variety, veracity, and value.

In this regard, big data management refers to the process of ``controlling flows of large volume, high velocity, heterogeneous and/or uncertain data to create value''. To ensure that the project is effectively focused on value creation, business and IT alignment is an essential principle. The presented Big Data Management Canvas consists of the following five phases:
1) Data preparation: combination of data from different sources into a single platform with consistent access for analytics
2) Data analytics: extraction of actionable knowledge directly from data through a process of discovery, hypothesis formulation and hypothesis testing.
3) Data interaction: definition of how users interfere with the data analysis results. 
4) Data effectuation: use of data analysis results to create value in products and services.
5) Data intelligence: ability of the organization to acquire and apply knowledge and skills in data management. In this sense, there are three types of knowledge and skills: a) knowledge generated from data b) knowledge about data and data management and c) knowledge and skills necessary for data analytics and management

This framework is based on an epistemic model of big data management as a cognitive system: it theorizes that knowledge emerges not by passive observation, but by ``iterative closed loops of purposely creating and observing changes in the environment''. Targeting this vision towards data analytics, it claims that the knowledge that big data analytics generates emerges from the interaction of data scientists and end-users with existing databases and analysis results. 

Overall, this methodology offers a different perspective of data science, explicitly prioritizing the value creation from data. Most approaches to big data research aim at data storage, computation and analytics instead of knowledge and value, that should be the result of big data processing. This change of perspective affects the project development, continuously balancing and aligning business and technology. 

\begin{itemize}
\item[$\color{green}\bigtriangleup$] Prioritizes value creation from data, rather than focusing on storage, computation and analytics.
\item[$\color{red}\bigtriangledown$] Team management challenges are not addressed
\end{itemize}

\begin{figure*}[bt!]
	\centering
	\begin{minipage}{0.24\textwidth}
		\includegraphics[width=\linewidth]{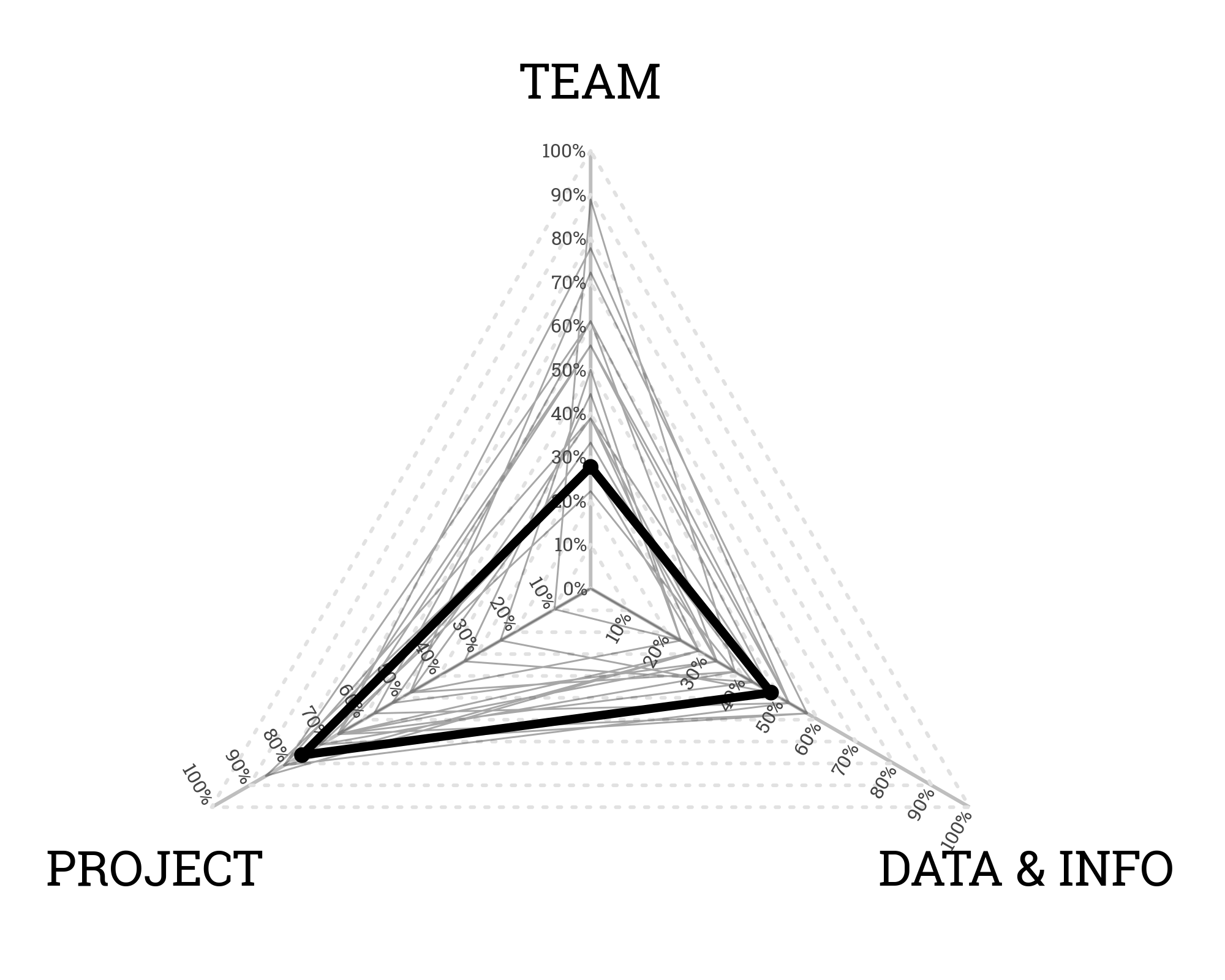}
    		\caption{Big Data Management Canvas} \label{bigdata_canvas}
	\end{minipage}\hfill
	\begin{minipage}{0.24\textwidth}
    		\includegraphics[width=\linewidth]{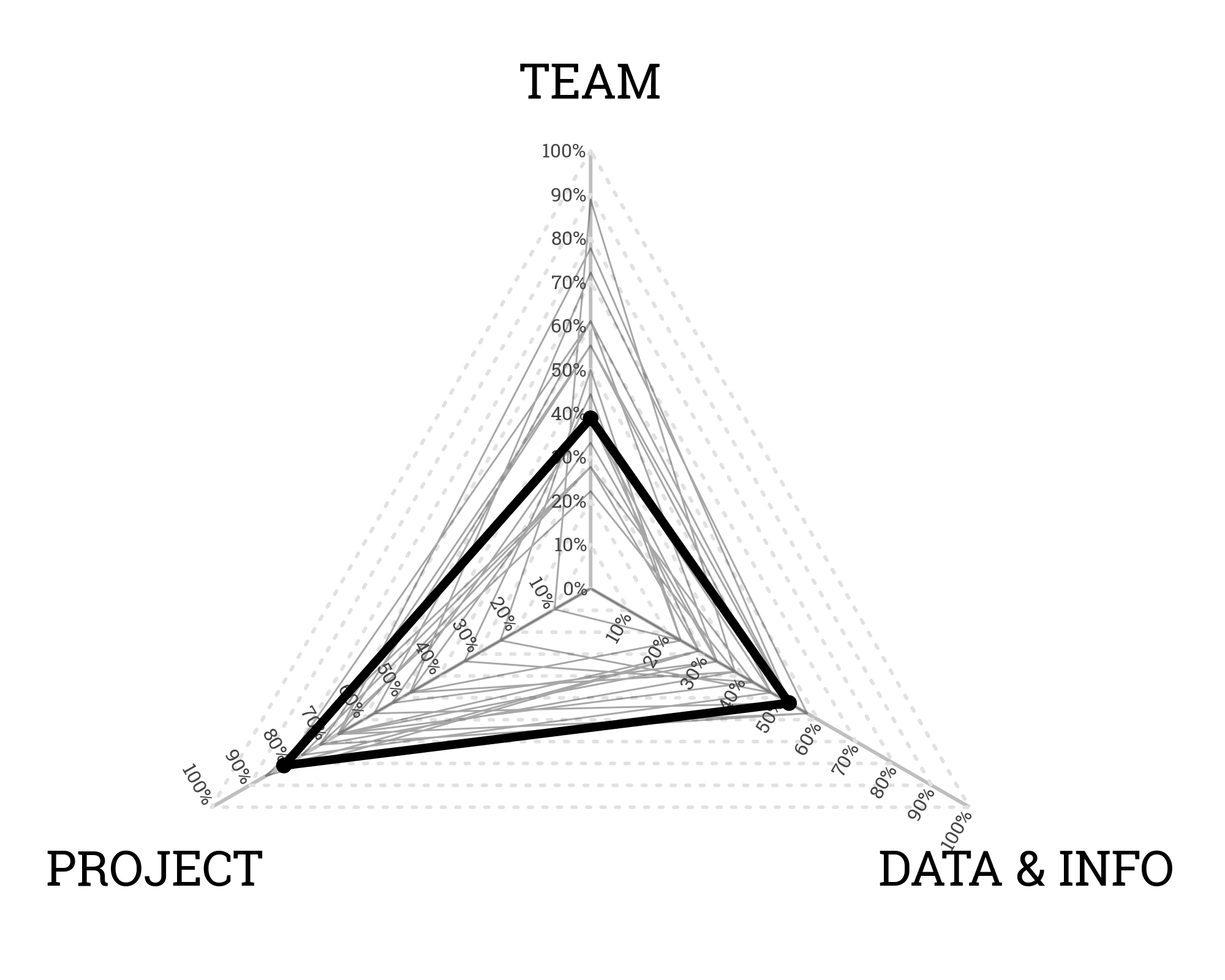}
		\caption{Agile Delivery Framework} \label{agile_delivery}
  	\end{minipage}\hfill
  	\begin{minipage}{0.24\textwidth}
    		\includegraphics[width=\linewidth]{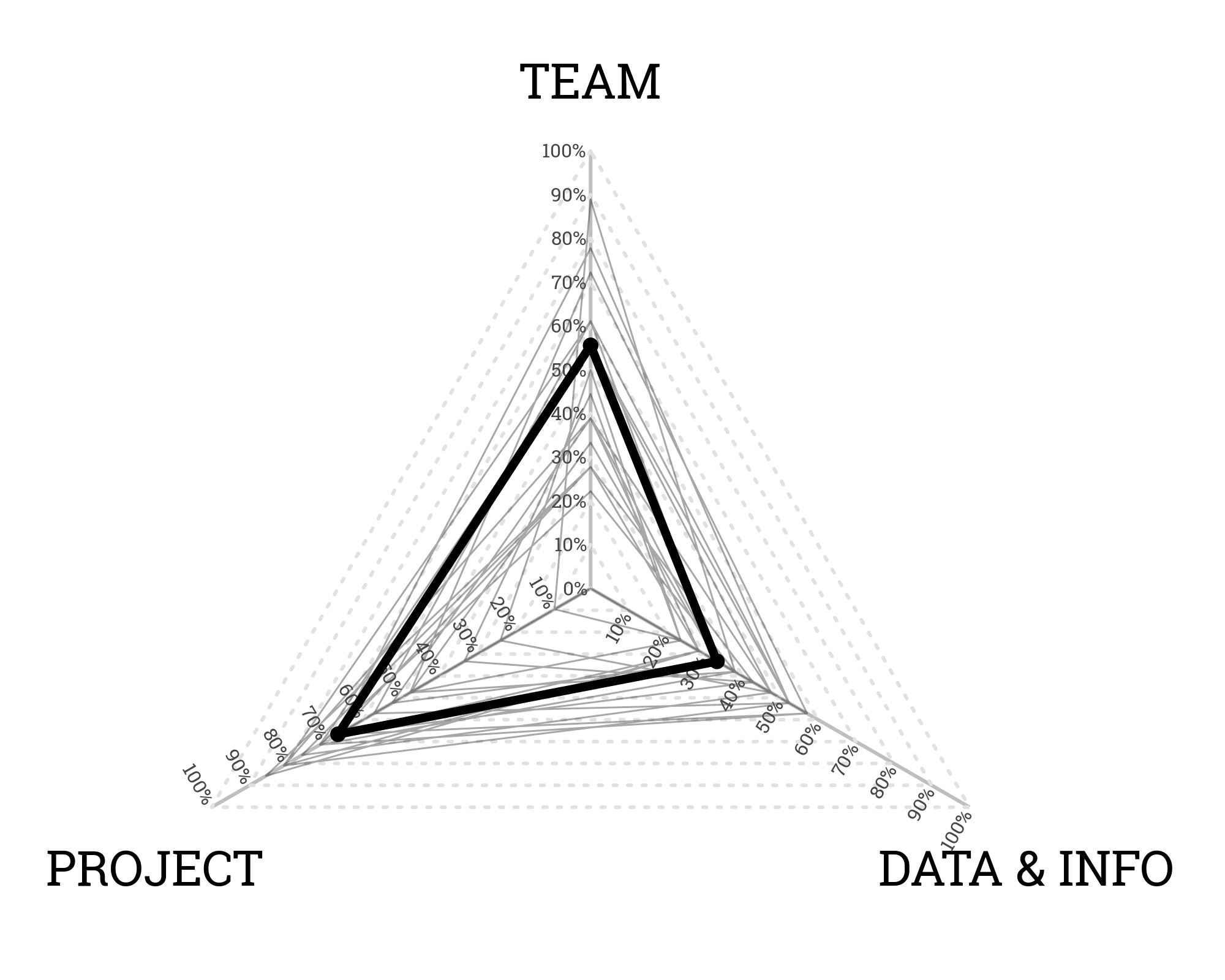}
    		\caption{Systematic Research on Big Data} \label{systematic}
  	\end{minipage}\hfill
  	\begin{minipage}{0.24\textwidth}
    		\includegraphics[width=\linewidth]{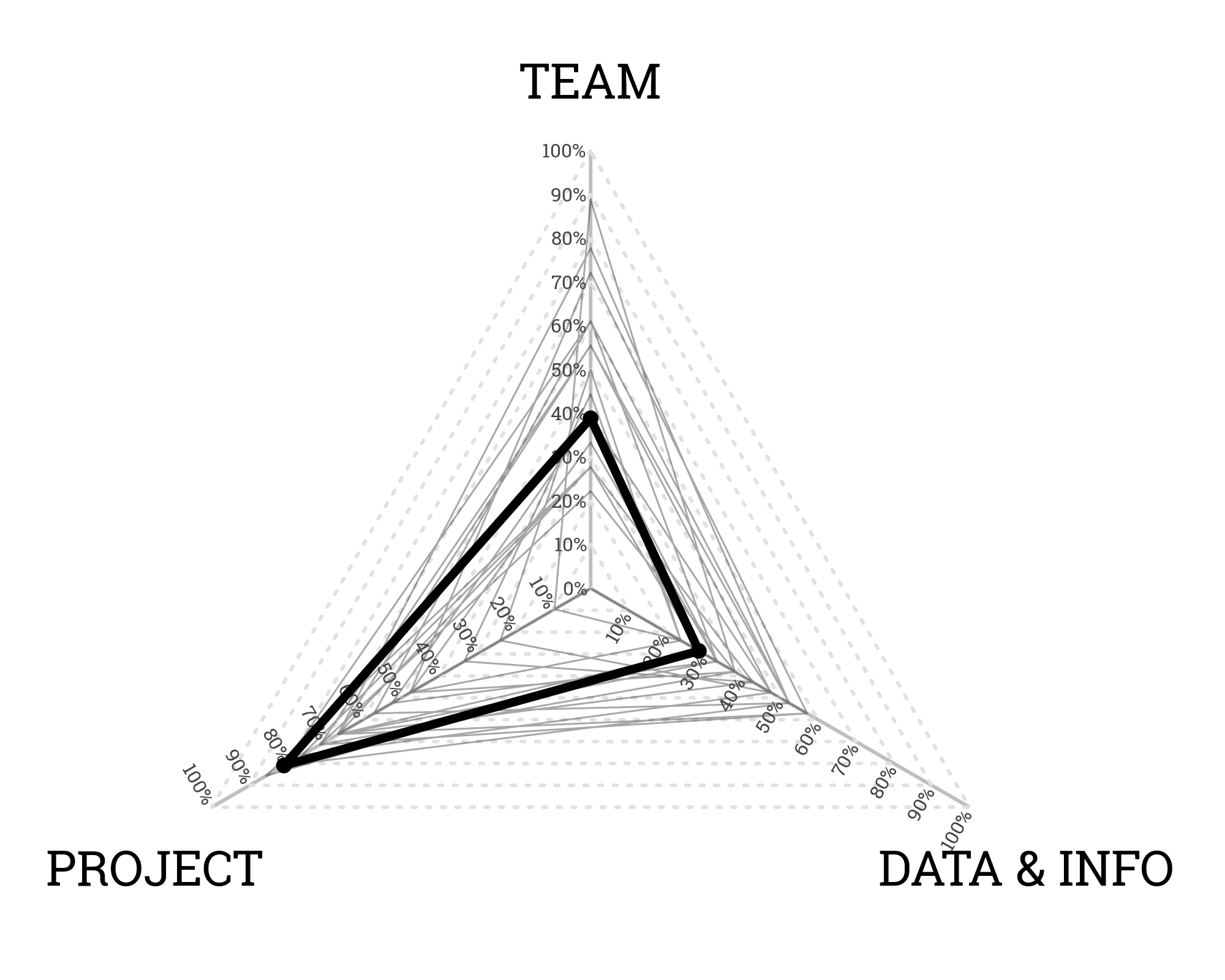}
    		\caption{Big Data Managing Framework} \label{bigdata_managing}
  	\end{minipage}
\end{figure*}

\subsection{Agile Delivery Framework} \label{sec:agile_delivery}

\noindent
Larson and Chang \cite{larson_agile} propose a framework based on the synthesis of agile principles with Business Intelligence (BI), fast analytics and data science. There are two layers of strategic tasks: (A) the top layer includes BI delivery and (B) the bottom layer includes fast analytics and data science. 

In the top layer there are five sequential steps: discovery, design, development, deployment and value delivery.
A1) ``discovery'' is where stakeholders determine the business requirements and define operating boundaries.
A2) ``design'' focuses on modeling and establishing the architecture of the system.
A3) ``development'' is a very broad phase that includes a wide array of activities, for instance, coding ETLs or scripting scheduling jobs.
A4) ``deployment'' focuses on integration of new functionalities and capabilities into production.
A5) ``value delivery'' includes maintenance, change management and end user feedback.

The bottom layer includes six sequential steps: scope, data acquisition, analysis, model development, validation and deployment.
B1) ``scope'' defines the problem statement and the scope of data sources.
B2) ``data acquisition'' acquires the data from the data lake and assesses the value of the data sources.
B3) ``analysis'' visualizes the data and creates a data profiling report.
B4) ``model development'' fits statistical and machine learning models to the data.
B5) ``validation'' ratifies the quality of the model. Both the model development and the validation are carried out via timeboxed iterations and agile methods.
B6) ``deployment'' prepares dashboards and other visualization tools.

The framework is designed to encourage successful business and IT stakeholder collaboration. For the authors, data science is inherently agile as the process is carried out through iterations, and data science teams are usually composed by small teams and require collaboration between business partners and technical experts. Similarly, the BI lifecycle has three phases in which the agile approach can be suitable: the discovery, design and development phases may benefit from iterative cycles and small time-boxed increments, even though there are not any software programming tasks. The main novel approach of this methodology is that completely separates the business intelligence and the data analysis worlds. In fact, it proposes two methodologies that evolve parallel to each other and by means of agile methods pledges an effective collaboration between these two parts.

\begin{itemize}
\item[$\color{green}\bigtriangleup$] Two methodologies for business intelligence and the data analysis that work in parallel
\item[$\color{red}\bigtriangledown$] Data management challenges and reproducibility issues are omitted
\end{itemize}

\subsection{Systematic Research on Big Data} \label{sec:systematic}

\noindent
Das et al. \cite{das_systematic_research} explore the systematization of data-driven research practices. The presented process is composed of eight Agile analytic steps, that start with a given dataset and end with the research output. In this sense, \cite{collier_agile} is the recommended development style for the process. 

First, information is extracted and cleaned. Even though the core of data-driven research is the data, the authors claim that the process should start from a question of what is intended to obtain from the research and the data. Therefore, the next phase is the preliminary data analysis, in which some patterns and information are revealed from the data. These patterns can be extracted with the help of unsupervised methods, and will be used to find an appropriate target and starting point of the research. Hence, the generated data patterns and discoveries lead to the definition of the research goal or research hypothesis. 

Once the setup of the research goal is defined, more work can be put on the extracted data. Otherwise, the research goal must be modified, and preliminary data analysis tasks such as descriptive analytics need to be carried out again. Therefore, further analysis can be done given the goal of the research, selecting the most relevant features and building machine learning models. The output from these models or predictive systems is further evaluated. Within an Agile process, the output evaluation can be done in iterative enhancements. Finally, it is important to communicate and report the results effectively, using meaningful infographics and data visualization techniques. These steps may be repeated in an iterative fashion until the desired results or level of performance is achieved.

This methodology suggests to execute the above workflow with a small subset of the dataset, identify issues early on, and only when satisfactory, expand to the entire dataset. In addition to the use of iterative Agile planning and execution, the authors claim that a generalized dataset and a standardized data processing can make data-driven research more systematic and consistent.

In general, this framework provides a process for performing systematic research on big data and is supported by Agile methodologies along the development process. Agile can promote frequent collaboration and a value-driven development in an iterative, incremental fashion. However, not enough effort is put by the authors on defining the roles and interactions of the team members across the different stages of the project. Overall, this methodology is closer to the research and academic world, and since data science finds itself between research activities and business applications, it is very valuable to have a methodology to bring out the best from at least, one of the worlds.

\begin{itemize}
\item[$\color{green}\bigtriangleup$] Research-focused and systematic approach
\item[$\color{red}\bigtriangledown$] Does not define roles and interactions between team members across the different stages of the project
\end{itemize}

\subsection{Big Data Managing Framework} \label{sec:bigdata_managing}

\noindent
Dutta et al. \cite{dutta_ramco} present a new framework for implementing big data analytics projects that combines the change management aspects of an IT project management framework with the data management aspects of an analytics framework.
This methodology presents a iterative and cyclical process for a big data project, which is organized into three distinct phases: strategic groundwork, data analytics and implementation. 

The strategic groundwork phase involves the identification of the business problem, brainstorming activities, the conceptualization of the solution to be adopted and the formation of the project teams. More in detail, the ``business problem'' phase sets the right expectations of the stakeholders, and dissipate any myths about what a big data project can achieve. The ``research'' phase studies how similar problems have been solved and it also looks for different analytics products available in the market. Then the team is conformed, and it recommends building it up with people from various backgrounds: business units, IT experts, data scientist, experts in the field, business decision makers, etc. Thus, this framework stands up for cross-functional and multidisciplinary teams. With all of this, the project roadmap is prepared, gathering up the major activities of the project, with timelines and designated people. This roadmap must be flexible enough, and should focus on execution and delivery rather than aiming to a strict adherence to the plan.

The data analytics phase has a clear structure, formed by ``data collection and examination'', ``data analysis and modeling'', ``data visualization'' and ``insight generation''. As observed in other methodologies, the use of a preliminary phase for data collection and exploration is widely used. Nevertheless, relying on data visualization techniques to present the results of data analysis tasks is a novel point, and indicates that the authors are very much concerned about getting all the stakeholders involved on the project development. Besides, innovative visual representation of the data can help in insight generation and in detecting similar and anomalous patterns in data. The data analytics phase is concluded with the insight generation step, which is another innovative point  occasionally omitted by other methodologies. It is important to understand the underlying latent reasons behind data trends. This last step takes the analysis to insights and possible actionable business input that can be of value to the organization.

Finally, during the implementation phase the solution is integrated with the IT system and the necessary people are trained to learn how to use it. The integration is a friction-prone phase, since it involves dealing with existing IT systems and designing a robust architecture is an challenging task. To make the implementation of the new system easier, users need to be ``trained in how to use the tools and the data that is available''. In this way, users will feel more comfortable with a new source of insights while taking their business decisions.

\begin{itemize}
\item[$\color{green}\bigtriangleup$] Emphasizes change management aspects, such as cross-functional team formation and training of people
\item[$\color{red}\bigtriangledown$] Not enough emphasis is given to the validation of data analytics and machine learning models 
\end{itemize}

\begin{figure*}[bt!]
	\centering
	\begin{minipage}{0.24\textwidth}
		\includegraphics[width=\linewidth]{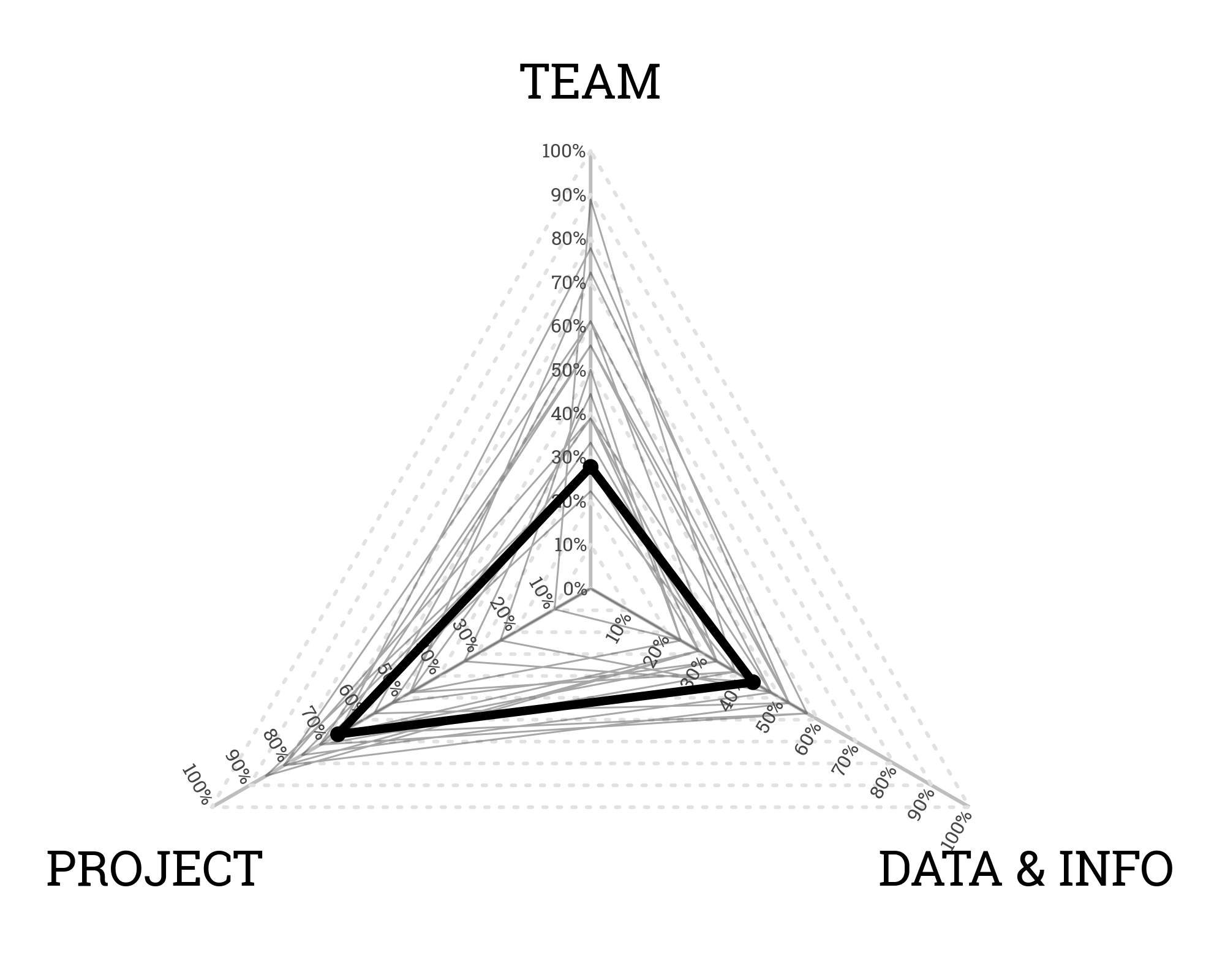}
		\caption{Data Science Edge} \label{dsedge}
	\end{minipage}\hfill
	\begin{minipage}{0.24\textwidth}
    		\includegraphics[width=\linewidth]{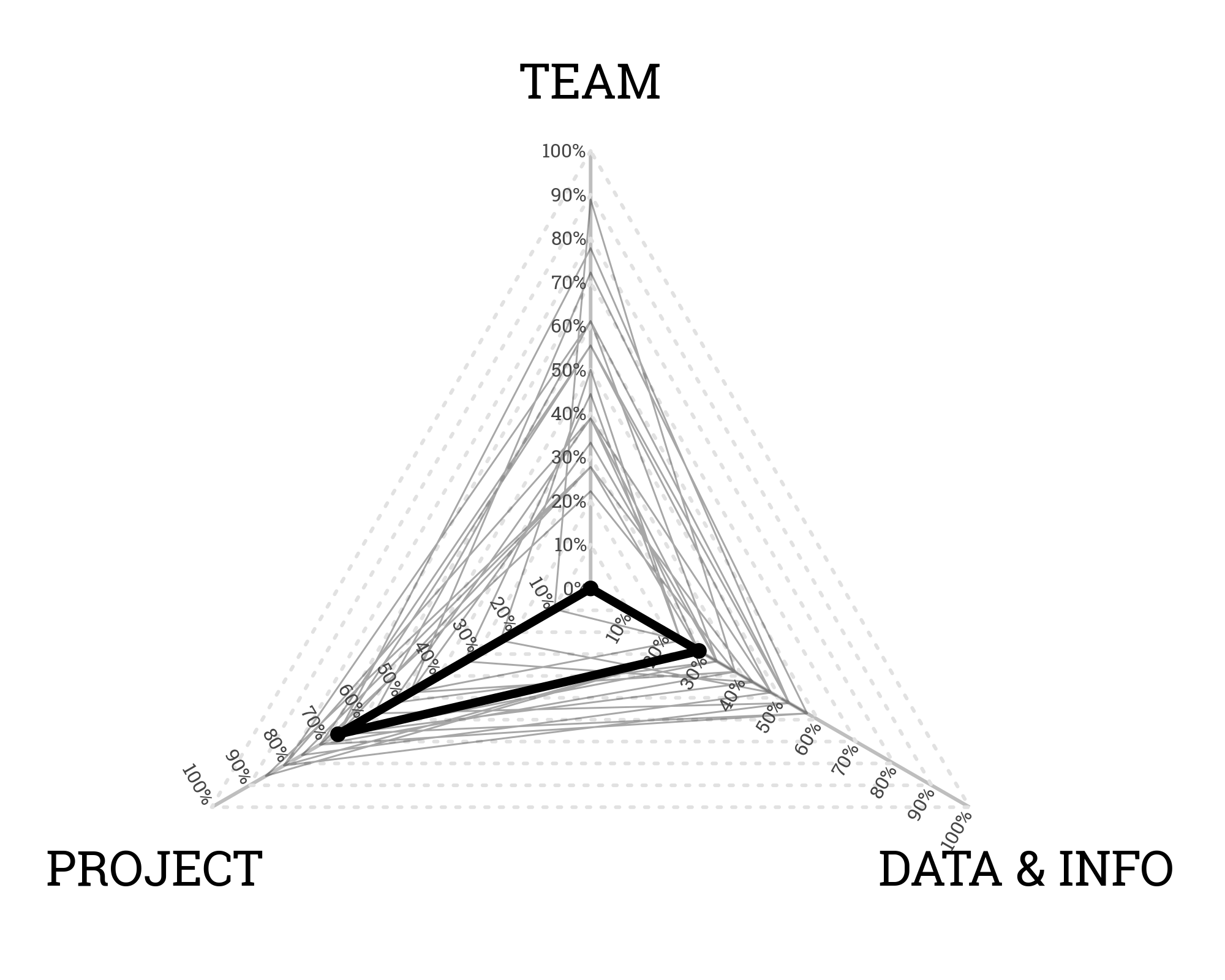}
    		\caption{FMDS} \label{fmds}
  	\end{minipage}\hfill
  	\begin{minipage}{0.24\textwidth}
    		\includegraphics[width=\linewidth]{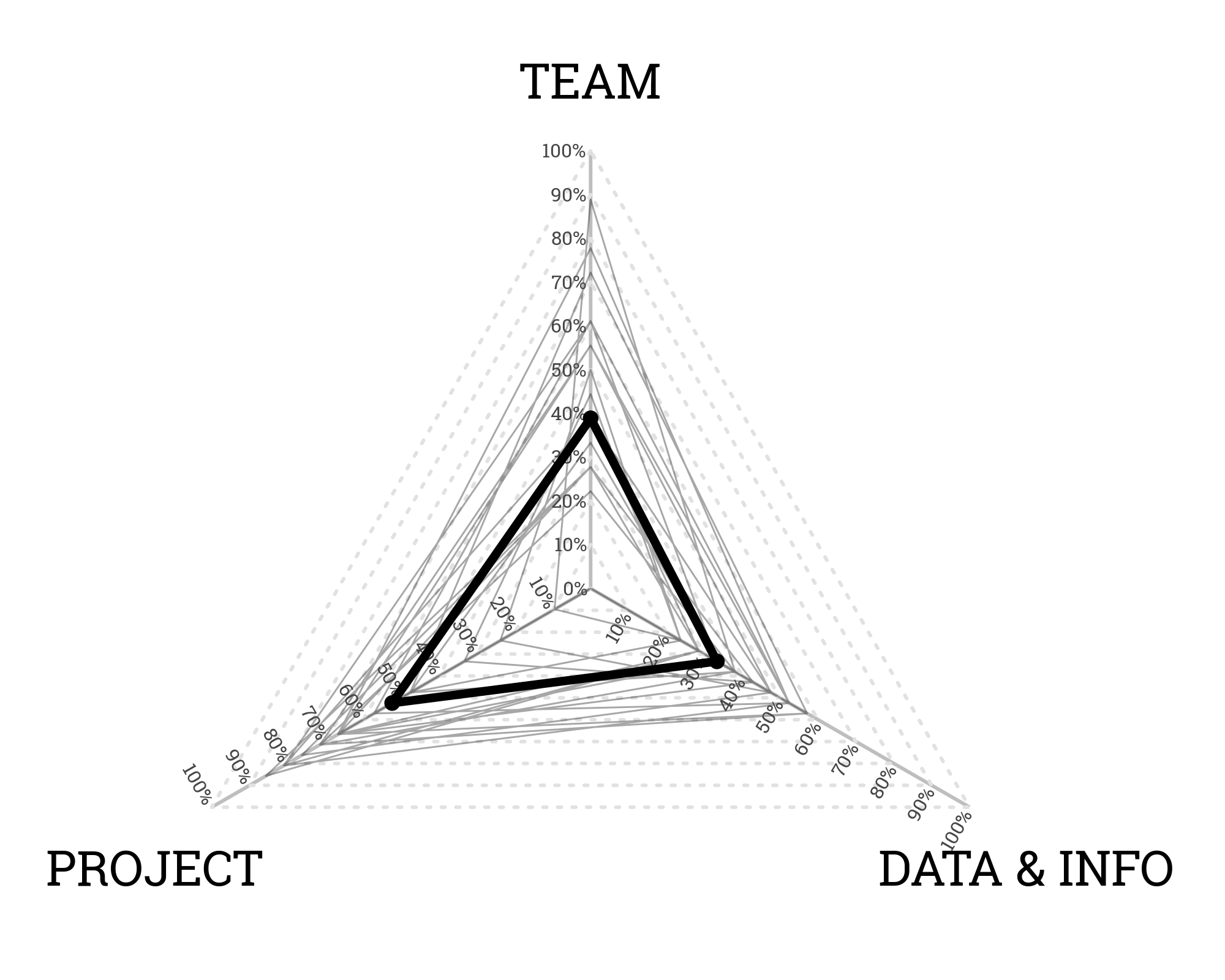}
    		\caption{Analytics Canvas} \label{analytics_canvas}
  	\end{minipage}\hfill
  	\begin{minipage}{0.24\textwidth}
    		\includegraphics[width=\linewidth]{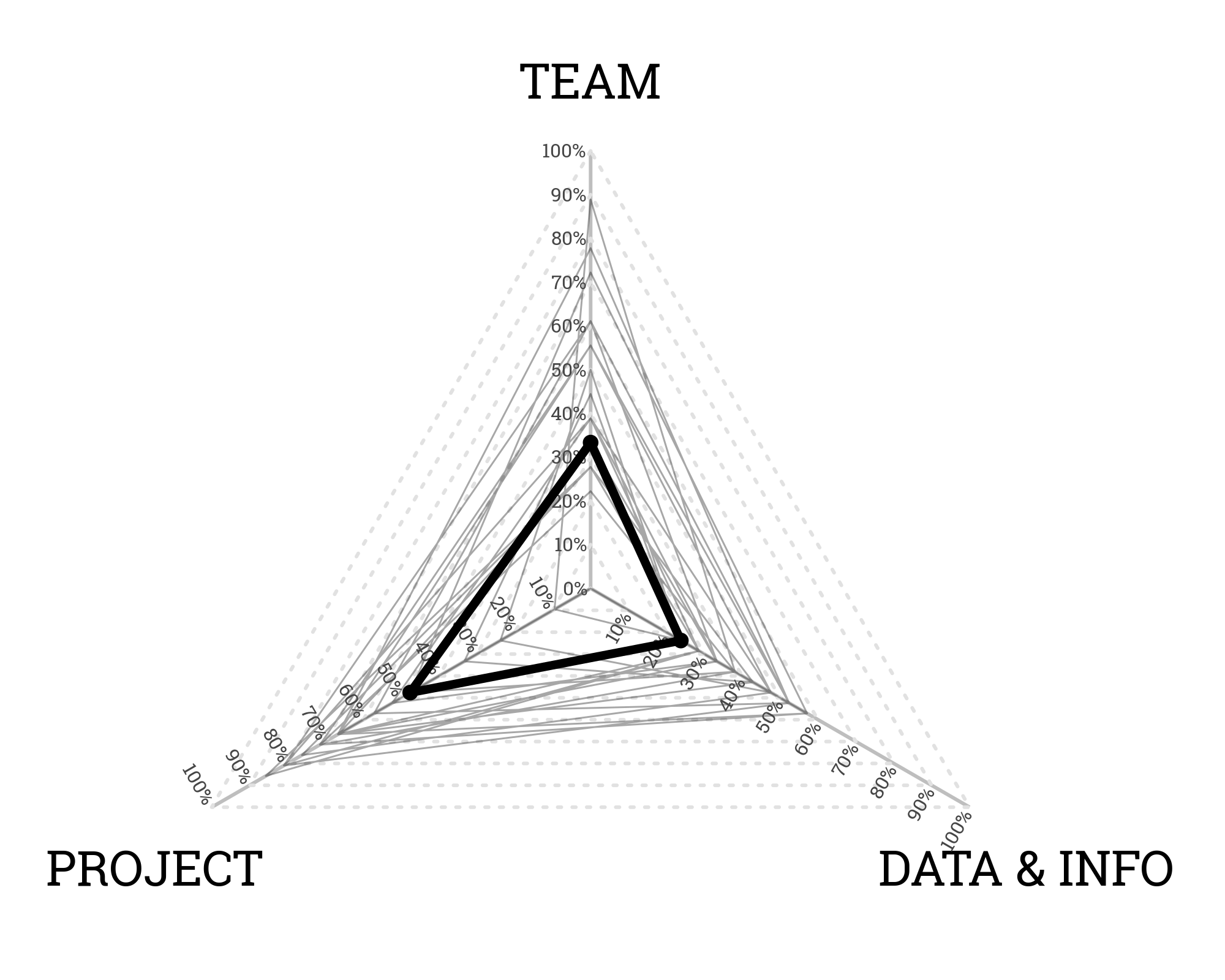}
		\caption{AI Ops} \label{aiops}
  	\end{minipage}
\end{figure*}

\subsection{Data Science Edge} \label{sec:dsedge}

\noindent
The Data Science Edge (DSE) methodology is introduced along two articles by Grady et al. \cite{grady_kdd, grady_agile}. It is a enhanced process model to accommodate big data technologies and data science activities. DSE provides a complete analytics lifecycle, with a five step model - plan, collect, curate, analyze, act, which are organized around the maturity of the data to address the project requirements. The authors divide the lifecycle into four quadrants - assess, architect, build and improve - which are detailed here below. 

The first quadrant - assess - consists in the planning, analysis of alternatives and rough order of magnitude estimation process. It includes the requirements and definition of any critical success factors. The activities within this first assessment phase correspond to the ``plan'' stage: defining organizational boundaries, justifying the investment and attending policy and governance concerns.

``Architect'' is the second quadrant and it consists on the translation of the requirements into a solution. The ``collect'' and ``curate'' stages are included on this phase. More specifically, the ``collect'' stage is responsible for the databases management, data distribution and dynamic retrieval. Likewise, during the ``curate'' stage the following activities are gathered: exploratory visualization, privacy and data fusion activities, and data quality assessment checks, among others.

The ``build'' quadrant comprises the ``analyze'' and ``act'' stages and consists of the development, test and deployment of the technical solution. Novel activities in the stage are: the search for simpler questions, latency and concurrency, or the correlation versus causation problem. The last quadrant, ``improve'' consists of the operation and management of the system as well as an analysis of ``innovative ways the system performance could be improved''.

On the whole, Grady et al. propose a novel big data analytics process model that extends CRISP-DM and that provides useful enhancements. It is claimed that DSE serves as a complete lifecycle to knowledge discovery, which includes data storage, collection and software development. The authors explain how the DSE process model aligns to agile methodologies and propose the required conceptual changes to adopt agile in data analytics. Finally, the authors conclude that by following the agile methodologies, the outcomes could be expected more quickly to make a decision and to validate the current state of the project.

\begin{itemize}
\item[$\color{green}\bigtriangleup$] Enhanced CRISP-DM process model to accommodate big data technologies and data science activities
\item[$\color{red}\bigtriangledown$] Challenges related to team management are omitted
\end{itemize}

\subsection{Foundational Methodology for Data Science} \label{sec:fmds}

\noindent
The Foundational Methodology for Data Science \cite{rollings_foundational_ibm} by IBM has some similarities with CRISP-DM, but it provides a number of new practices. FMDS's ten steps illustrate an iterative nature of the process of data science, with several phases joined together by closed loops. The ten stages are self-explanatory: 1) business understanding 2) analytic approach 3) data requirements 4) data collection 5) data understanding 6) data preparation 7) modeling 8) evaluation 9) deployment 10) feedback. 

It should be pointed out the interesting position of the analytic approach phase at the beginning of the project, just after understanding the business and without having any data collection nor exploratory analysis. Even though author's point of view is understood, in reality data scientists have difficulties to chose the analytic approach prior to the exploring the data. Framing the business problem in the context of statistical and machine learning techniques usually requires a preliminary study using descriptive statistics and visualization techniques. Overall this methodology structures a data science project in more phases (10) than CRISP-DM (6), but has the same drawbacks, in terms of lack a definition of the different roles of the team, and zero concerns on reproducibility, knowledge accumulation and data security.

\begin{itemize}
\item[$\color{green}\bigtriangleup$] Provides new practices and extends CRISP-DM process model
\item[$\color{red}\bigtriangledown$] Inherits some of the drawbacks of CRISP-DM, especially on team and data management
\end{itemize}

\subsection{Analytics Canvas} \label{sec:analytics_canvas}

\noindent
The Analytics Canvas by Arno Kuhn \cite{kuhn_analytics_canvas} is a semi-formal specification technique for the conceptual design of data analytics projects. It describes an analytics use case and documents the necessary data infrastructure during the early planning of an analytics project. It proposes a four layer model which differentiates between analytic use case, data analysis, data pools and data sources. 

The first step is understanding the domain and identifying the analytics use case: root-cause analysis, process monitoring, predictive maintenance or process control. Management agents usually initiate this stage, as they have a wide vision of the company. Based on the analytics use case, data sources are specified by domain experts: sensors, control systems, ERP systems, CRM, etc. In addition to knowing where the data comes from, it is necessary to specify the places where data is stored. For this purpose, the IT expert is the assigned person. Finally, the analytics use case is linked to a corresponding data analysis task: descriptive, diagnostic, predictive or prescriptive. Here the data scientist is the main agent involved.

Therefore, the Analytics Canvas assigns a specialized role to each phase: management, domain expert, IT expert, data scientist. The team is also supervised by the analytics architect. This canvas is very useful to structure the project in its early stages and identify its main constructs. Incorporating such a tool helps setting up clear objectives and promotes a transparent communication between stakeholders. Apart from describing the analytics use case and the necessary data infrastructure, the Analytics Canvas allows clear description and differentiation of the roles of stakeholders, and thus enables interdisciplinary communication and collaboration.

\begin{itemize}
\item[$\color{green}\bigtriangleup$] Helps on the conceptual design of data analytics projects, primarily during the early phases
\item[$\color{red}\bigtriangledown$] Hard to implement as an scalable framework along the entire project development
\end{itemize}

\subsection{AI Ops} \label{sec:aiops}

\noindent
John Thomas presents a systematic approach to ``operationalizing'' data science \cite{thomas_ai_ops}, which covers managing the complete end-to-end lifecycle of data science. The author referrers to these set of considerations as AI Ops, which include: a) Scope b) Understand c) Build (dev) d) Deploy and run (QA) and e) Deploy, run, manage (Prod)

For each stage, AI-Ops defines the necessary roles: data scientist, business user, data steward, data provider, data consumer, data engineer, software engineer, AI operations, etc. which helps organizing the project and improves coordination. However, there are no guidelines for how these roles collaborate and communicate with each other.

During the scope stage, the methodology insists on having clear business KPIs. Proceeding without clear KPIs may result in the team evaluating the project success by the model performance, rather than by its impact on the business. In this sense, correlating model performance metrics and trust/transparency elements to the business KPIs is indeed a frequent challenge. Without this information, it is difficult for the business to get a view of whether a project is succeeding. 

The data understanding phase is pivotal: AI-Ops proposes to establish the appropriate rules and policies to govern access to data. For instance, it empowers data science teams to shop for data in a central catalog: first exploring and understanding data, after which they can request a data feed and exploit it on the build phase. 

This methodology splits the deployment stage into two separated steps: quality assessment (QA) and production (prod), in order to make sure that any model in production meets the business requirements and the quality standards as well. Overall this methodology is focused on the operationalization of the project, thus giving more importance to the deployment, the investment on IT infrastructure and continuous development and integration stages.

\begin{itemize}
\item[$\color{green}\bigtriangleup$] Primarily focused on the deployment and operationalization of the project
\item[$\color{red}\bigtriangledown$] Does not offer guidelines for how different roles collaborate and communicate with each other \\
Reproducibility and knowledge retaining issues are left unexplored during model building phase
\end{itemize}

\subsection{Data Science Workflow} \label{sec:dsworkflow}

\noindent
Data Science Workflow \cite{guo_ds_workflow} by Philip Guo introduces a modern research programming workflow. There are four main phases: preparation of the data, alternating between running the analysis and reflection to interpret the outputs, and finally dissemination of results.

In the preparation phase data scientists acquire the data and clean it. The acquisition can be challenging in terms of data management, data provenance and data storage, but the definition of this preliminary phase is quite straightforward. From there follows the core activity of data science, that is, the analysis phase. Here Guo presents a separate iterative process in which programming scripts are prepared and executed. The outputs of these scripts are inspected and after a debugging task, the scripts are further edited. Guo claims that the faster the data scientist can make it through each iteration, the more insights could potentially be obtained per unit time.

Whereas the analysis phase involves programming, the reflection phase involves thinking and communicating about the outputs of analyses. After inspecting a set of output files, a data scientist might take notes, hold meetings and make comparisons. The insights obtained from this reflection phase are used to explore new alternatives by adjusting script code and execution parameters.

The final phase is disseminating the results, most commonly in the form of written reports. The challenge here is to take all the various notes, sketches, scripts and output data files created throughout the process to aid in writing. Some data scientists also distribute their software so that other researchers can reproduce their experiments.

Overall, discerning the analysis from the reflection phase is the main strength of this workflow by Guo. The exploratory nature of data science projects makes challenging to establish a waterfall execution and development. Hence, usually it is needed to go back and forth to find the appropriate insights and the optimal model for the business. Separating the pure programming tasks (analysis) from the discussion of the results is very valuable for data scientists, that can get lost within the programming spiral. 

\begin{itemize}
\item[$\color{green}\bigtriangleup$] Discerns the analysis phase from the reflection phase 
\item[$\color{red}\bigtriangledown$] Explicitly focused on research/science, not immediately applicable in business case
\end{itemize} 

\begin{figure*}[bt!]
	\centering
	\begin{minipage}{0.24\textwidth}
		\includegraphics[width=\linewidth]{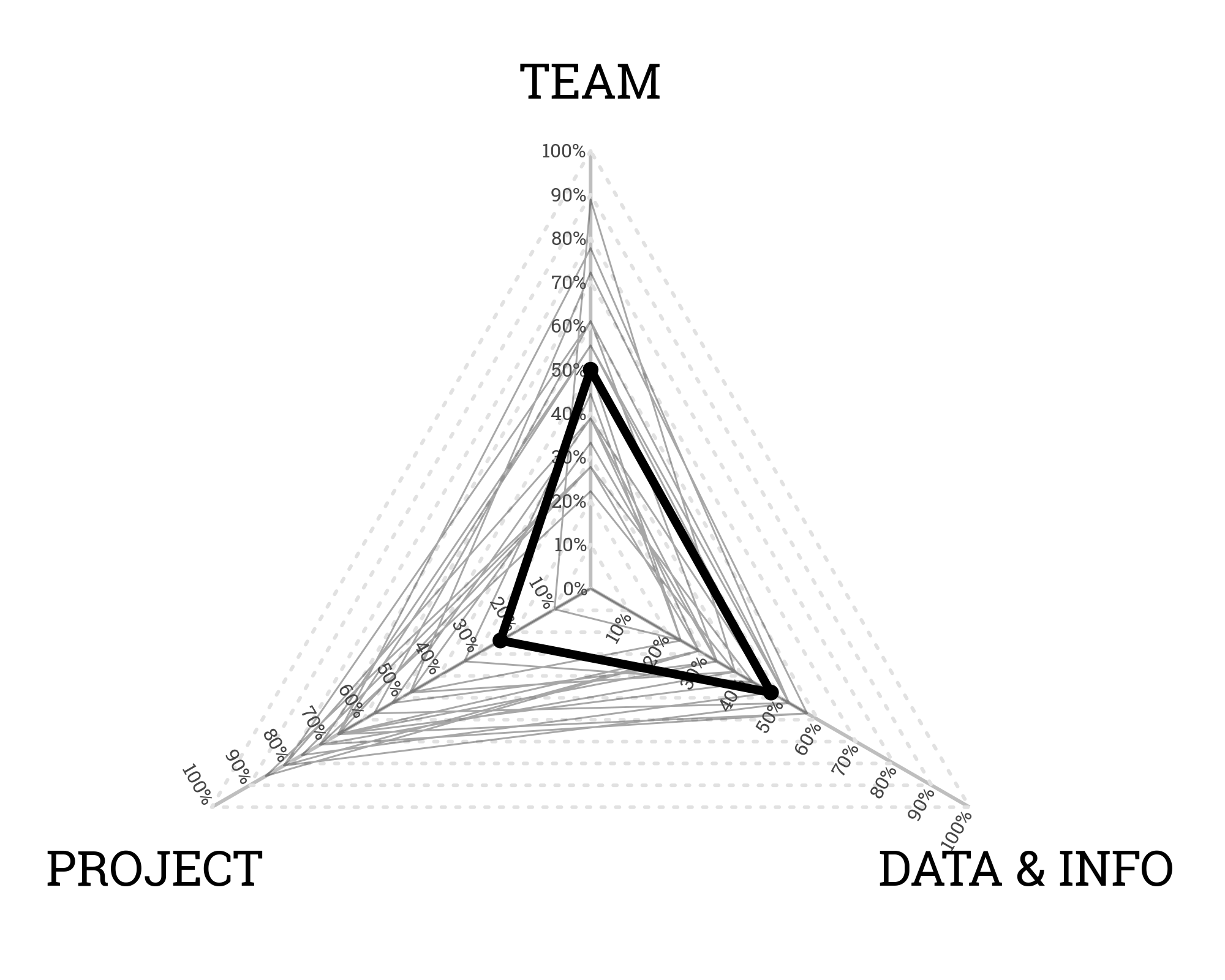}
    		\caption{Data Science Workflow} \label{dsworkflow}
	\end{minipage}\hfill
	\begin{minipage}{0.24\textwidth}
    		\includegraphics[width=\linewidth]{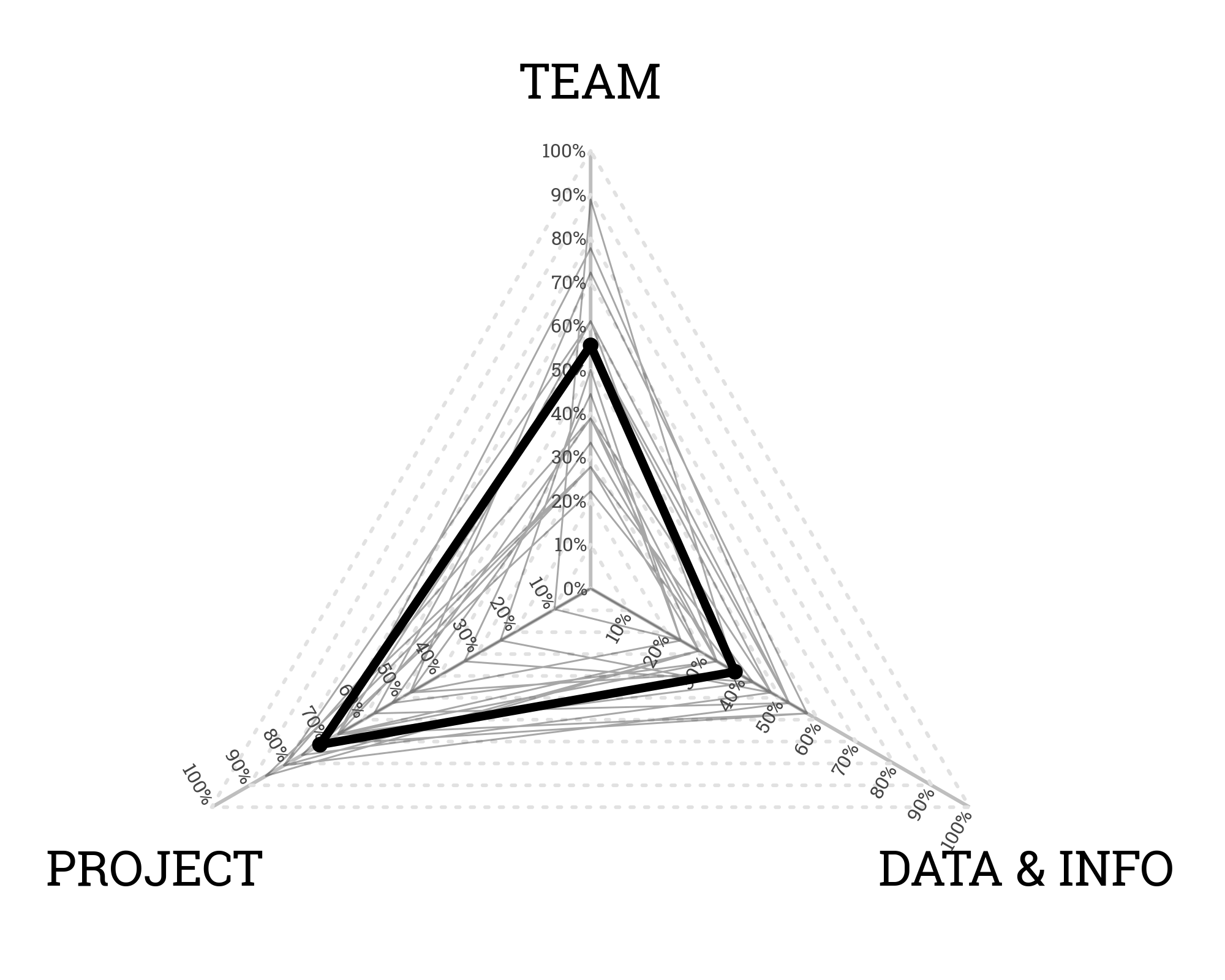}
    		\caption{EMC Data Analytics Lifecycle} \label{emc}
  	\end{minipage}\hfill
  	\begin{minipage}{0.24\textwidth}
    		\includegraphics[width=\linewidth]{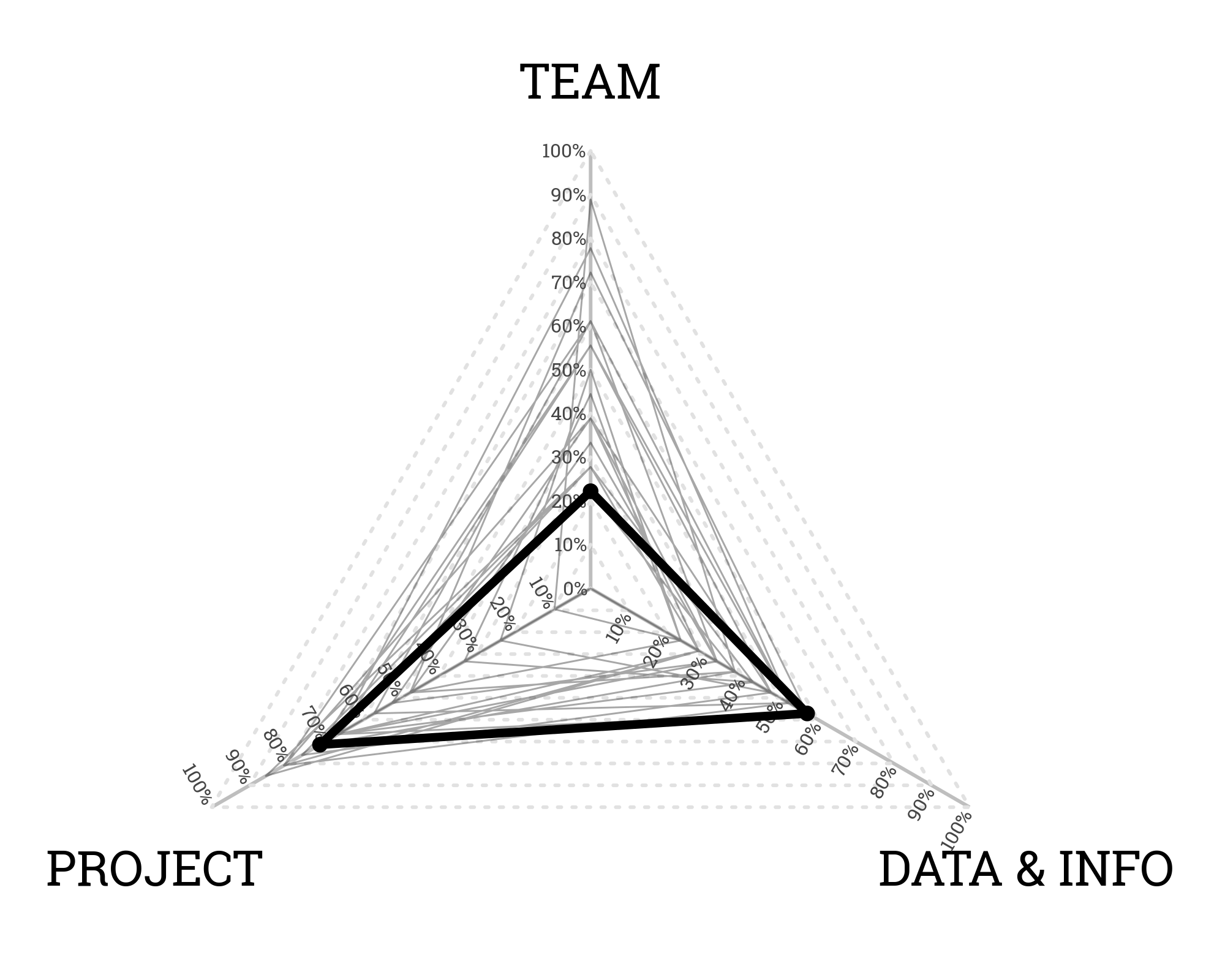}
    		\caption{Toward data mining engineering} \label{dm_engineering}
  	\end{minipage}
\end{figure*}

\subsection{EMC Data Analytics Lifecycle} \label{sec:emc}

\noindent
EMC Data Analytics Lifecycle by EMC \cite{dietrich_emc} is a framework for data science projects that was designed to convey several key points: A) Data science projects are iterative
B) Continuously test whether the team has accomplished enough to move forward
C) Focus work both up front, and at the end of the projects

This methodology defines the key roles for a successful analytics project: business user, project sponsor, project manager, business intelligence analyst, database admin, data engineer, data scientist. EMC also encompasses six iterative phases, which are shown in a circle figure: data discovery, data preparation, model planning, model building, communication of the results, and operationalization. 

During the discovery phase, the business problem is framed as an analytics challenge and the initial hypothesis are formulated. The team also assesses the resources available (people, technology and data). Once enough information is available to draft an analytic plan, data is transformed for further analysis on the data preparation phase.

EMC separates the model planning from the model building phase. During the model planning, the team determines the methods, techniques, and workflow it intends to follow for the subsequent model building phase. The team explores the data to learn about the relationships between variables and subsequently selects key variables and the most suitable models. Isolating the model planning from the building phase is a smart move, since it assists the iterative process of finding the optimal model. The back and forth process of obtaining the best model can be chaotic and confusing: a new experiment can take weeks to prepare, program and evaluate. Having a clear separation between model planning and model building sure enough helps during that stage. 

Presented as a separated phase, the communication of the results is a distinctive feature of this methodology. In this phase, the team, in collaboration with major stakeholders, determines if the results of the project are a success or a failure based on the criteria developed in the discovery phase. The team should identify key findings, quantify the business value, and develop a narrative to summarize and convey findings to stakeholders. Finally, on the operationalization phase, the team delivers final reports, briefings, code, and technical documents.

Overall, the EMC methodology provides a clear guideline for the data science lifecycle, and a definition of the key roles. It is very concerned about teams excessively focusing on phases two through four (data preparation, model planning and model building), and explicitly prevents them from jumping into doing modeling work before they are ready. Even though it establishes the key roles in a data science project and how they should collaborate and coordinate, it does not get into detail on how teams should communicate more effectively.

\begin{itemize}
\item[$\color{green}\bigtriangleup$] Prevents data scientists from jumping prematurely into modeling work
\item[$\color{red}\bigtriangledown$] Lacks a reproducibility and knowledge management setup
\end{itemize} 

\subsection{Toward data mining engineering} \label{sec:dm_engineering}

Marb\'an et al. \cite{marban_data_mining} propose to reuse ideas and concepts from software engineering model processes to redefine and add to the CRISP-DM process, and make it a data mining engineering standard.

The authors propose a standard that includes all the activities in a well-organized manner, describing the process phase by phase. Here only the distinctive points are discussed. The activities missing from CRISP-DM are primarily project management processes, integral processes and organizational processes. The project management processes establish the project structure and coordinate the project resources throughout the project lifecycle. CRISP-DM only takes into account the project plan, which is a small part of the project management. This methodology includes the lifecycle selection, and more processes that will not be further explained: acquisition, supply, initiation, project planning and project monitoring and control.

The integral processes are necessary to successfully complete the project activities. In this category the processes of evaluation, configuration management, documentation and user training are included. The organizational processes help to achieve a more effective organization, and it is recommended to adapt the SPICE standard. This group includes the processes of improvement (gather best practices, methods and tools), infrastructure (build best environments) and training.

\begin{figure*}[hbt!]
	\centering
	\includegraphics[width=1\linewidth]{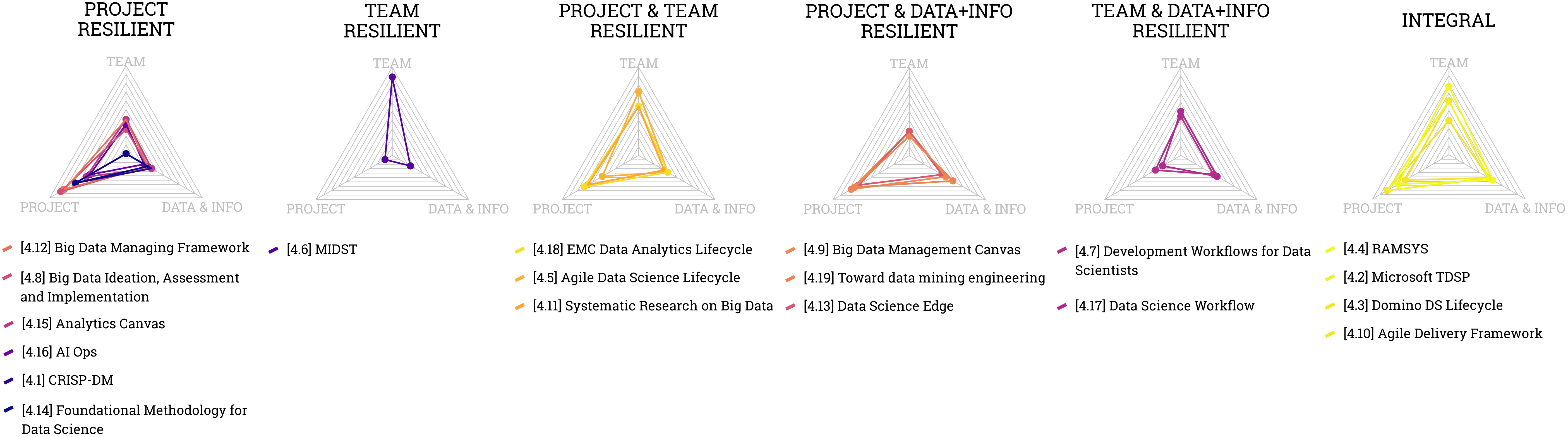}
	\caption{Taxonomy of methodologies for data science projects}
	\label{per_category}
\end{figure*}

The core processes in data science engineering projects are the development processes which are divided into pre-, KDD-, and post-development processes. The proposed engineering process model is very complete and almost covers all the dimensions for a successful project execution. It is based on very well studied and designed engineering standards and as a result are very suitable for large corporations and data science teams. Nevertheless, excessive effort is put into the secondary, supporting activities (organizational, integral and project processes) rather than on the core development activities. This extra undertaking may produce non-desired effects, burdening the execution of the project and creating inefficiencies due to processes complexity. In general terms, this methodology is very powerful in project management and information management aspects, as it conveys full control over every aspect of the project.

\begin{itemize}
\item[$\color{green}\bigtriangleup$] Thorough engineering process model for a successful project execution
\item[$\color{red}\bigtriangledown$] Team related challenges are left unexplored
\end{itemize}

\section{Discussion} \label{discussion}

So far, the challenges that arise when executing data science projects have been gathered on \cref{theoretical_framework}, and a critical review of methodologies has been conducted as summarized on \cref{results}. Based on the reviewed methodologies a taxonomy of methodologies is proposed, as illustrated on \cref{per_category}

The criteria for such taxonomy has focused on summarizing the categories each methodology adequately covers. For example, the EMC data analytics lifecycle \cite{dietrich_emc} methodology is resilient on project and team management challenges, whereas MIDST \cite{crowston_midst} clearly leans toward team management. This taxonomy is meant to better understand the types of methodologies that are available for executing data science projects.

Apart from the lack of a methodology usage in real data science projects that was mentioned on  \cref{theoretical_framework}, there is an absence of complete or integral methodologies across the literature. \textbf{In this sense, of the 19 reviewed methodologies, only 4 are classified as ``integral''}. Among these ``integral'' methodologies, certain aspects could be improved: TSDP \cite{microsoft_tdsp} has a strong dependence on Microsoft tools and technologies and their roles stick too closely to Microsoft services. For instance, TDSP's data scientist role is limited to the cloning of several repositories and to merely executing the data science project. The data scientist role should be broken down to more tangible roles, and detail their responsibilities outside the Microsoft universe. 

DominoLab lifecycle \cite{domino_data_lab} lacks tests to check the limitations and quality of data, and even though it integrates a team-based approach, it does not describe how teams should work to communicate, coordinate and collaborate effectively. RAMSYS \cite{moyle_ramsys} not completely address reproducibility and traceability assurance, and the investment on IT resources is not fully taken into account.Lastly, Agile Data Science Lifecycle \cite{jurney_agile_ds} is not overly concerned about the limitations of Machine Learning techniques, data security and privacy, nor does propose methods for results evaluation. 

\begingroup	
\renewcommand\arraystretch{1.25}	
\begin{table*}[ht]
	\centering
	\resizebox{1\textwidth}{!}{
	\begin{tabular}{C{7.3cm}lC{5cm}lC{8cm}}
		\cellcolor[HTML]{C0C0C0} PROJECT MANAGEMENT &   & \cellcolor[HTML]{C0C0C0}TEAM MANAGEMENT &   & \cellcolor[HTML]{C0C0C0}DATA \& INFORMATION MANAGEMENT \\ \cline{1-1} \cline{3-3} \cline{5-5} 
		\begin{tabular}[c]{@{}C{7.3cm}@{}} Definition of data science lifecycle workflow \\[4pt] Standardization of folder structure \\[4pt] Continuous project documentation \\[4pt] Visualization of project status \\[4pt] Alignment of data science \& business goals \\[4pt] Consolidation of performance \& success metrics \\[4pt] Separation of prototype \& product \end{tabular} & 
		&
		\begin{tabular}[c]{@{}C{5cm}@{}} Promotion \& communication of scientific findings \\[4pt] Role definition to aid coordination between team members and across stakeholders \\[4pt] Boost team collaboration: git workflow \& coding agreements \end{tabular} &
		&
		\begin{tabular}[c]{@{}C{8cm}@{}} \\[1pt] Reproducibility: creation of knowledge repository \\[4pt] Robust deployment: versioning code, data \& models \\[4pt] Model creation for knowledge and value generation \\[4pt] Traceability reinforcement: put journey over target \end{tabular} \\ \cline{1-1} \cline{3-3} \cline{5-5} 
	\end{tabular}
	}
	\caption{Summary of the proposed principles of integral methodologies for data science projects}
	\label{tab:summary}
\end{table*}
\endgroup

Besides, other methodologies which focus on project management tend to forget about the team executing the project, and often leave the management of data \& information behind. What is needed is a management methodology that takes into account the various data-centric needs of data science while also keeping in mind the application-focused uses of the models and other artifacts produced during an data science lifecycle, as proposed on \cite{walch_agile}.
In view of the reviewed methodologies, we think that, as every data science project has its own characteristics and it is hard to cover all the possibilities, it would be recommendable to establish the foundation for building an integral methodology. Consequently, we propose a conceptual framework that takes in the general features that a integral methodology for managing data science project could have. A framework that could cover all the pitfalls and issues listed above once provided with a set of processes and best practices from industry, business and academic projects. This framework can be used by other researchers as a roadmap to expand currently used methodologies or to design new ones. In fact, anyone with the vision for a new methodology could take this boiler-template and build its own methodology.

In this sense, we claim that an efficient data science methodology should not rely on project management methodologies alone, nor should be solely based on team management methodologies. In order to offer a complete solution for executing data science projects, three areas should be covered, as it is illustrated on \cref{proposal}:

\begin{figure}[hbt!]
	\centering
	\includegraphics[width=0.8\linewidth]{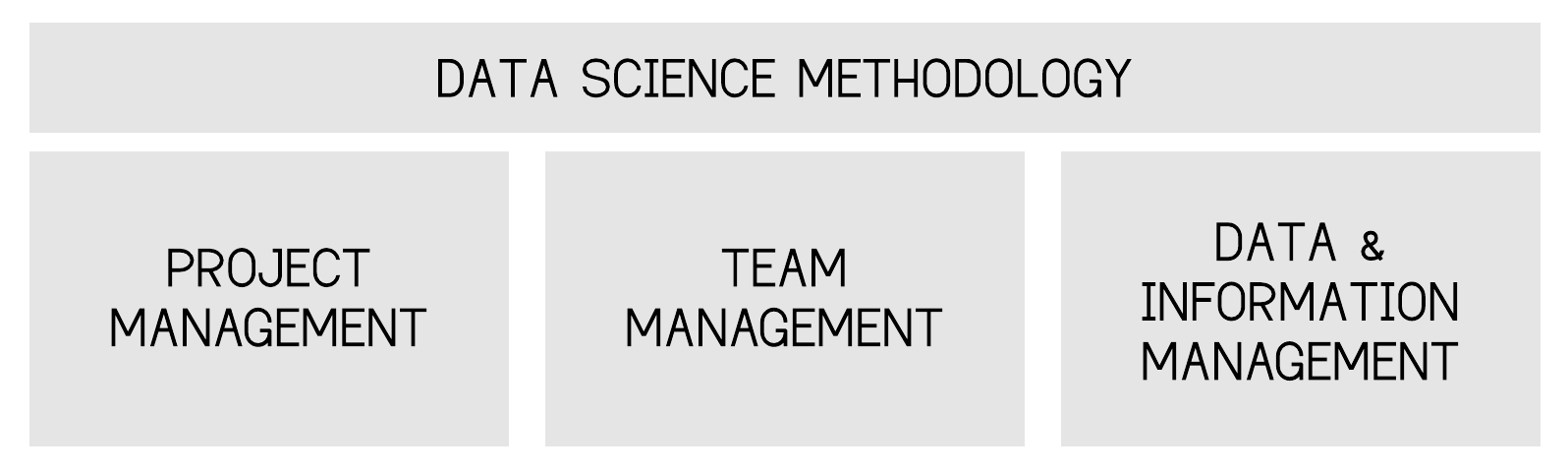}
	\caption{Proposed foundation stones of integral methodologies for data science projects}
	\label{proposal}
\end{figure}

\textbf{Project methodologies} are generally task-focused kind of methodologies that offer a guideline with the steps to follow, sometimes presented as a diagram-flow. Basically they try to define the data science lifecycle, which in most cases is very similar between different named methodologies \cite{foroughi_ds_methodologies}: business problem understanding, data gathering, data modeling, evaluation, implementation. Thus, their main goal is to prepare the main stages of a project so that it can be successfully completed. 

It seems true that data science projects are very hard to manage due to the uncertain nature of data among other factors, and thus their project failure rate is very elevated. One of the most critical points is that at the beginning of any project, data scientists are not familiar with the data, and thus it is hard to know the data quality and its potential to accomplish certain business objectives. This complicates a lot the project definition and the establishment of SMART objectives. Therefore, having an effective project management methodology is fundamental for the success of any project, and specially for data science projects, since they can: a) monitor in which stage the project is at any moment b) setup the project goals, its stages, outcomes of the project, its deliverables and c) manage the task scopes, and deliverables deadlines.

Regarding team management, data science is no more considered an individual task undertaken by "unicorn" data scientists. It is definitely a team effort, which requires a methodology that guides the way the team members communicate, coordinate and collaborate. That is why we need to add another dimension to the data science methodology, one that can help managing the role and responsibilities of each team member and helps also efficiently communicating the team progress. That is the objective of the so-called \textbf{team management methodologies}. The important point in this regard is that no project management methodology alone can ensure the success of the data science project. It is necessary to establish a teamwork methodology that coordinates and manages the tasks to be performed and their priority.

We could say that these two dimensions, project and team management could be applied to almost any project. Of course, the methodologies for each case would need to be adapted and adjusted. However, data science projects inherently work with data, and usually with lots of data, to extract knowledge that can answer some specific questions of a business, a machine, a system o an agent, thus generating value from the data. Therefore, we believe it is fundamental to have a general approach that guides the generation of insights and knowledge from data, so called \textbf{data \& information management}.

The frontier between information and knowledge can be delicate, but in words of \cite{terra_information_knowledge}, information is described as refined data, whereas knowledge is useful information. Based on the DIKW pyramid, from data we can extract information, from information knowledge and from knowledge wisdom. We claim that data science projects are not eager to extract wisdom from data, but to extract useful information, thus knowledge, from data.

Therefore, it is necessary to manage data insights and communicate the experiments outcomes in a way that contributes to the team and project knowledge. In the field of knowledge management, this means that tacit knowledge must be continuously be converted into explicit knowledge. In order to make this conversion from tacit knowledge, which is intangible, to explicit knowledge, which can be transported as goods, information and a communication channel is required. Therefore, the information management and the team communication processes are completely related.

Information is usually extracted from data to gain knowledge, and with that knowledge decisions are made. Depending on the problem in hand and the quality and potential use of the information, more or less human input is necessary to make decisions and further take actions. The primary strategic goal of the analytics project (descriptive, diagnostic, predictive or prescriptive) may affect the type of information that needs to be managed, and as a consequence also may influence the importance of the information management stage. Each strategic goal tries to answer different questions and also define the type of information that is extracted from data.

With only descriptive information from the data, it is required to control and manage the human knowledge under use; whereas with more complex models (diagnostic and predictive) it is much more important to manage the data, the model, inputs, outputs, etc. Therefore, with more useful information extracted from the data, the path to decision is shortened. 
    
Besides, \cite{ferraris_knowledge_management} points out that knowledge management will be the key source for competitive advantage for companies. With data science growing with more algorithms, more infrastructure, the ability to capture and exploit unique insights will be a key differentiator. As the tendency stands today, in few years time data science and the application of machine learning models will become more of a commodity. During this transformation data science will become less about framework, machine learning and statistics expertise and more about data management and successfully articulating business needs, which includes knowledge management.

All things considered, \cref{tab:summary} summarizes the main foundations for our framework: the principles are laid down so that further research can be developed to design the processes and the adequate tools.

\section{Conclusions} \label{conclusions}

Overall, in this article a conceptual framework is proposed for designing integral methodologies for the management of data science projects. In this regard the three foundation stones for such methodologies have been presented: project, team and data \& information management. It is important to point out that this framework must be constantly evolving and improving to adapt to new challenges in data science.

The proposed framework is built upon a critical review of currently available data science methodologies, from which a taxonomy of methodologies has been developed. This taxonomy is based on a quantitative evaluation of how each methodology overcomes the challenges presented on  \cref{theoretical_framework}. In this regard, the obtained scores were estimated by the authors' respective research groups. Even though this evaluation may contain some bias, we believe it gives an initial estimation of the strengths and weaknesses of each methodology. In this sense, we would like to extend the presented analysis to experts and researchers and also to practitioners in the field.

\section*{Acknowledgements}
This research did not receive any specific grant from funding agencies in the public, commercial, or not-for-profit sectors.

\appendix
\section{Triangular area} \label{appendixA}

This supplement contains the procedure to calculate the area of the triangle taking as origin the first Fermat point. The area is a relevant measure for the integrity of each methodology, as the distances from the first Fermat point represent their percentage score on project, team and information management.

Given the coordinates of the three vertices $A,B,C$ of any triangle, the area of the triangle is given by:

\begin{equation} \label{eq:A1}
	\Delta = \left | \frac{ A_{x}(B_{y}-C_{y}) + B_{x}(C_{y}-A_{y}) + C_{x}(A_{y}-B_{y}) }{2} \right |
\end{equation}

\begin{figure}[ht]
	\centering
	\includegraphics[width=0.7\linewidth]{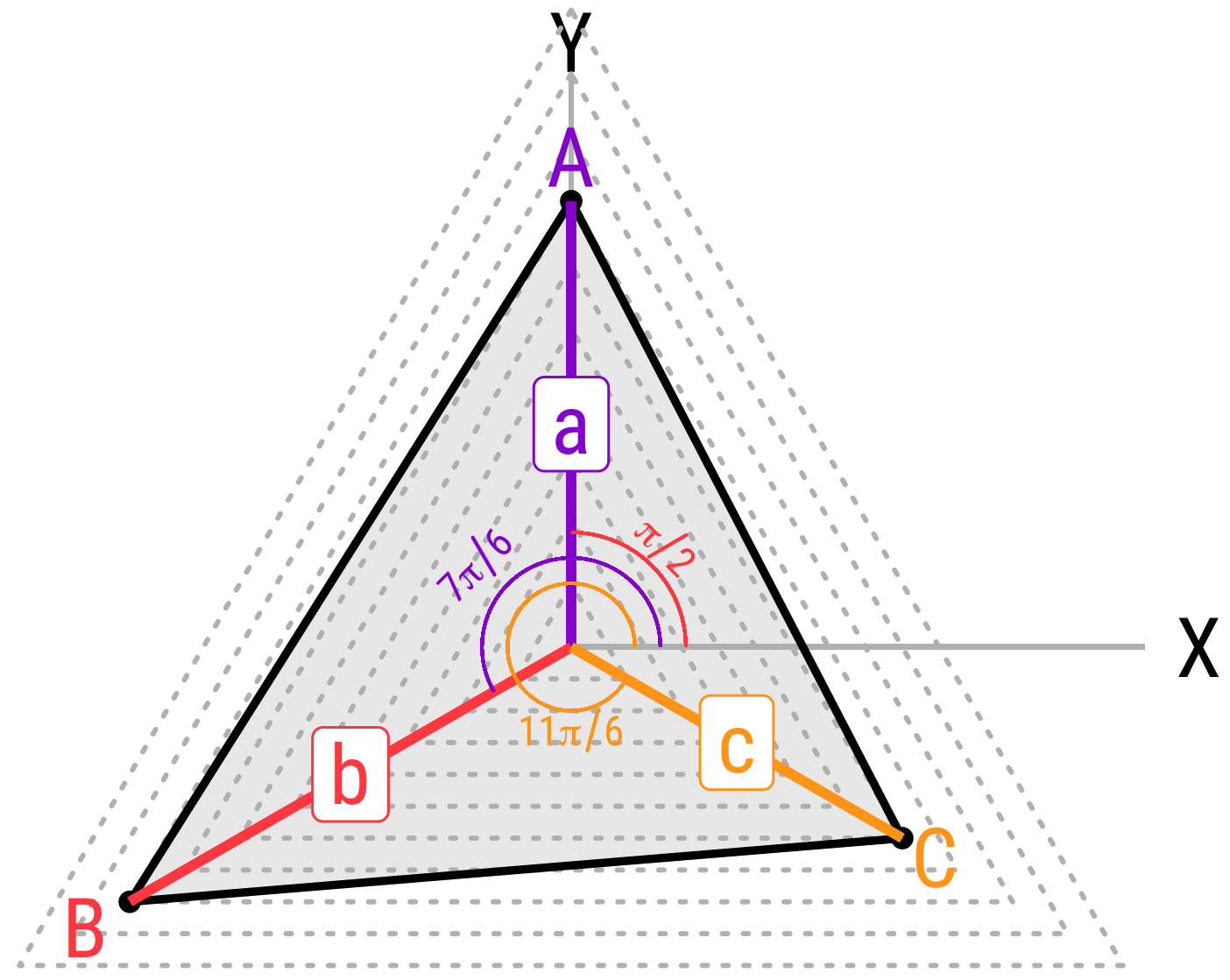}
	\caption{Triangular diagram: origin of coordinates, triangle vertices and corresponding angles}
	\label{triangle}
\end{figure}

With the origin of coordinates as the first Fermat point of the triangle, the coordinates of the vertices $A,B,C$ are defined as:
\begin{equation}
	\begin{array}{l}
		A_{x} = a \cdot cos(\pi/2) ;\quad A_{y} = a \cdot sin(\pi/2);     \\
		B_{x} = b \cdot cos(7\pi/6) ;\quad B_{y} = b \cdot sin(7\pi/6);   \\
		C_{x} = c \cdot cos(11\pi/6) ;\quad C_{y} = c \cdot sin(11\pi/6); 
	\end{array}
\end{equation}

Then, inserting these expressions for $A,B,C$ coordinates into \ref{eq:A1} yields the final expression for the area of the triangle. To reach this simplification it must be taken into account that $cos(7\pi/6) = -cos(11\pi/6) = \sqrt{3}/2$ and also $sin(7\pi/6) = sin(11\pi/6) = -1/2$.

\begin{equation}
	\Delta = \left | \frac{\sqrt{3}}{4}\left ( a \cdot b +  a \cdot c + b \cdot c \right ) \right |
\end{equation}

\bibliography{mybib}

\end{document}